\documentclass[10pt,twocolumn,letterpaper]{article}
\usepackage{iccv}
\usepackage{times}
\usepackage{epsfig}
\usepackage{graphicx}
\usepackage{amsmath}
\usepackage{amssymb}
\usepackage{booktabs}
\usepackage{times}
\usepackage{mathtools}
\usepackage{amsthm}
\usepackage{microtype}
\usepackage{subfig}
\usepackage{enumitem}
\usepackage{mmstyle}
\usepackage{algorithm}
\usepackage{bm}
\usepackage{algorithmic}
\usepackage{float}
\usepackage{listings}
\usepackage{makecell}  % \Xhline
\usepackage{diagbox} % diagonal table cell
\usepackage{multirow}
\usepackage{bbm}
\usepackage[misc]{ifsym}
\usepackage[pagebackref=true,breaklinks=true,colorlinks,bookmarks=false]{hyperref}

\usepackage[table]{xcolor}

\newcommand{\tablestyle}[2]{\setlength{\tabcolsep}{#1}\renewcommand{\arraystretch}{#2}\centering\footnotesize}
\newlength\savewidth\newcommand\shline{\noalign{\global\savewidth\arrayrulewidth
  \global\arrayrulewidth 1pt}\hline\noalign{\global\arrayrulewidth\savewidth}}
  
\newcolumntype{x}[1]{>{\centering\arraybackslash}p{#1pt}}
\newcolumntype{y}[1]{>{\raggedright\arraybackslash}p{#1pt}}
\newcolumntype{z}[1]{>{\raggedleft\arraybackslash}p{#1pt}}
\renewcommand{\paragraph}[1]{\vspace{1.25mm}\noindent\textbf{#1}}
\definecolor{deemph}{gray}{0.6}

\usepackage[capitalize]{cleveref}
\crefname{section}{Sec.}{Secs.}
\Crefname{section}{Section}{Sections}
\crefname{table}{Tab.}{Tabs.}
\Crefname{table}{Table}{Tables}

\def\iccvPaperID{9856} % *** Enter the ICCV Paper ID here

\newcommand\blfootnote[1]{%	
  \begingroup
  \renewcommand\thefootnote{}\footnote{#1}%
  \addtocounter{footnote}{-1}%
  \endgroup
}

\begin{document}

%%%%%%%%% TITLE
\title{Deep Equilibrium Object Detection}
\author{Shuai Wang\textsuperscript{1} \quad \quad Yao Teng\textsuperscript{1} \quad \quad  Limin Wang\textsuperscript{1,2,~\Letter}\\
$^1$State Key Laboratory for Novel Software Technology, Nanjing University \quad $^2$Shanghai AI Lab \\ [0.2cm]
{\bf \url{https://github.com/MCG-NJU/DEQDet}} 
}
\maketitle

\pagestyle{empty}
\thispagestyle{empty}

% !TEX root = ../main.tex

\begin{abstract}

Query-based object detectors directly decode image features into object instances with a set of learnable queries. These query vectors are progressively refined to stable meaningful representations through a sequence of decoder layers, and then used to directly predict object locations and categories with simple FFN heads. 
In this paper, we present a new query-based object detector (DEQDet) by designing a deep equilibrium decoder. Our DEQ decoder models the query vector refinement as the fixed point solving of an {\em \bf implicit} layer and is equivalent to applying {\em \bf infinite} steps of refinement.
To be more specific to object decoding, we use a two-step unrolled equilibrium equation to explicitly capture the query vector refinement. Accordingly, we are able to incorporate refinement awareness into the DEQ training with the inexact gradient back-propagation (RAG). In addition, to stabilize the training of our DEQDet and improve its generalization ability, we devise the deep supervision scheme on the optimization path of DEQ with refinement-aware perturbation~(RAP).
Our experiments demonstrate DEQDet converges faster, consumes less memory, and achieves better results than the baseline counterpart (AdaMixer). In particular, our DEQDet with ResNet50 backbone and 300 queries achieves the $49.5$ \textit{mAP} and $33.0$ \textit{AP$_s$} on the MS COCO benchmark under $2\times$ training scheme (24 epochs).

\end{abstract}

\blfootnote{\Letter: Corresponding author (lmwang@nju.edu.cn).}
% !TEX root = ../main.tex

\section{Introduction}
\label{sec:introduction}

\begin{figure}
    \centering
    \vskip 0.2in
    \subfloat[FFN view on query-based object detector.~\label{fig:ffn_view}]{
        \includegraphics[width=0.95\linewidth]{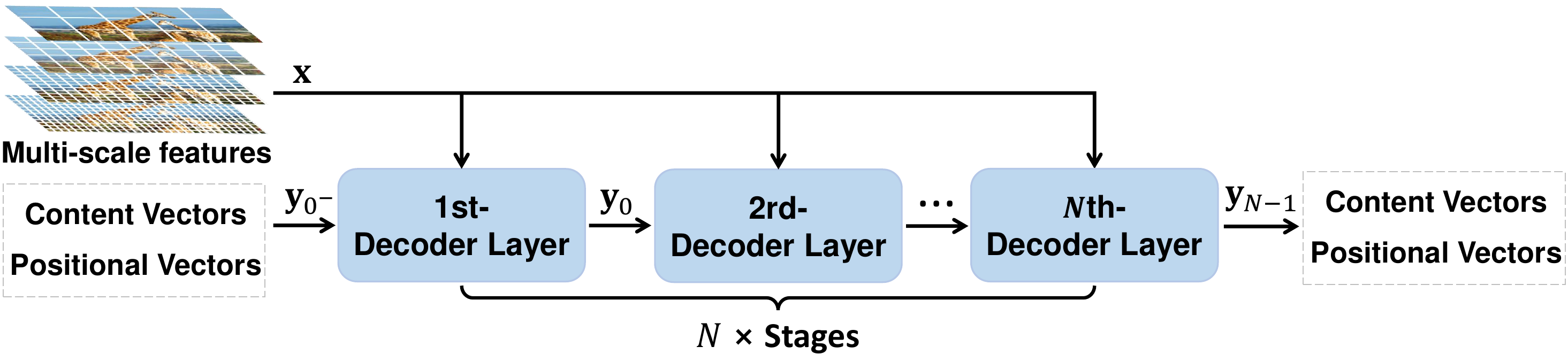}
    }\\ \vspace{-1mm}
    \subfloat[RNN view on query-based object detector.~\label{fig:rnn_view}]{
        \includegraphics[width=0.95\linewidth]{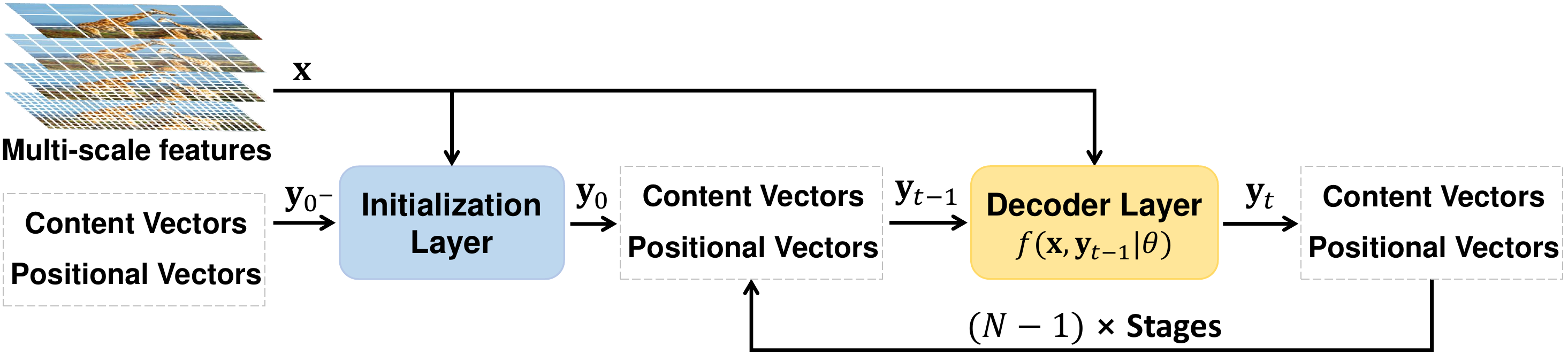}
    }\\ \vspace{-1mm}
    \subfloat[Our DEQDet view on query-based object detector.~\label{fig:eq_view}]{
        \includegraphics[width=0.95\linewidth]{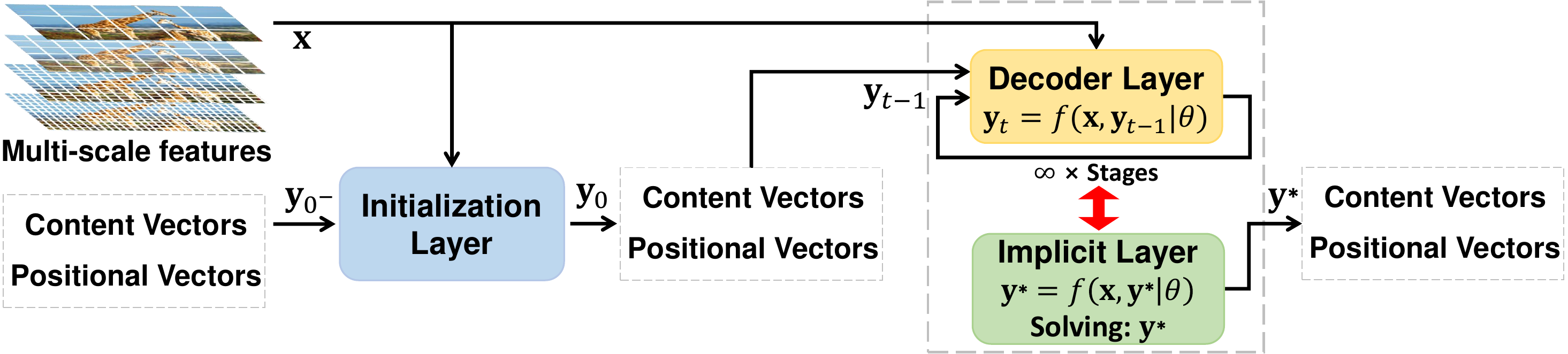}
    }\\
    \caption{\textbf{Different views on query-based object detector.}~~(a)~In FFN view, the decoder consists of stacked non-shared decoder layers \eg AdaMixer~\cite{adamixer} (b)~In RNN view, the decoder consists of weight-tied decoder layers. (c)~In our DEQDet, decoder performs refinement in a RNN manner, but model it as the fixed point solving of an implicit layer with infinite steps. The red arrow means equivalence.}
    \label{fig:intro}
    \vskip -0.1in
\end{figure}

Object detection~\cite{fasterrcnn,focalloss,detr,sparsercnn,adamixer} is a fundamental task in computer vision research.
Its purpose is to identify the locations and categories of all object instances in an image.
This task is challenging because object detector is usually required to deal with large variations in object instances. 
Traditional object detectors often make dense predictions based on a large quantity of candidates such as anchor boxes~\cite{ssd,fasterrcnn,cascadercnn} or reference points~\cite{duan2019centernet,zhou2021centernet2,cornernet}.
Among these models, one-stage object detectors~\cite{ssd, fcos,focalloss,centernet} directly classify the candidates and regress the bounding boxes based on them.
In addition, two-stage detectors~\cite{fasterrcnn,fastrcnn} adopt an additional initialization step to select a set of coarse object proposals from dense anchors, and then \textit{refine} their locations and predict their categories. These two types of detectors often require hand-crafted post-processing techniques such as NMS to yield the final detection results.

Recently, query-based object detectors~\cite{detr,sparsercnn,adamixer} present a new paradigm for object detection. 
As shown in ~\cref{fig:ffn_view}, the detectors are composed of a set of learnable object queries and several stacked non-shared decoder layers (e.g. cross-attention~\cite{detr}, dynamic convolution~\cite{sparsercnn}, dynamic MLPMixer~\cite{adamixer}).
In such a detector, these query vectors are progressively refined by each decoder layer, where image features are sampled or attended and transformed based on each query.
After several steps of refinements, these query vectors could be directly transformed into object predictions with a simple FFN head.
The success of query-based object detectors yields a flexible and simple paradigm of directly decoding object instances from images without any dense assumption (e.g. dense anchors) or post processing.

Despite the great success achieved by query-based object detectors, some important issues on their design still remain. 
First, {\em parameter efficiency is an important issue for these detectors}~\cite{sparsercnn,adamixer}. Each decoder layer performs the same task of query refinement but has its own parameters, which leads to the large numbers of parameters and makes them prone to overfitting. 
Second, {\em depth of refinement (decoder layers) is another critical factor in detector design.} Intuitively, increasing refinement depth would scale up the detector capacity and hopefully contribute to a better detection performance with a proper optimization method. 
To address these critical issues, we first come up with a new RNN perspective on the query vector update as shown in~\cref{fig:rnn_view}, partially inspired by RNN-based optical flow estimation~\cite{teed2020raft}.
In this sense, we employ the same transformation in each decoder layer (known as weight tying), which provides a parameter-efficient way to scale up these query-based object detectors and could be viewed as a kind of regularization technique. Yet, this recurrent query refinement would yield significant computational and memory overhead due to the tracking of long hidden-state history in the BPTT algorithm~\cite{bptt}. In addition, it still needs to determine the number of decoder layers. Therefore, as shown in~\cref{fig:eq_view}, we further improve this weight-tying refinement to an extreme version, and model it as the fixed point solving process of an implicit layer with an efficient deep equilibrium model (DEQ)~\cite{deq}. This DEQ view on query-based object detection is able to simultaneously reduce its model parameters and increase its refinement depth to an infinite level.

Specifically, in this paper, we present a new query-based object detector, termed as {\bf DEQDet}, by designing a deep equilibrium decoder. 
To be more specific, it is based on the recent research of implicit models like DEQ~\cite{deq} and the recent query-based detector of AdaMixer~\cite{adamixer}.
Different from the previous query-based detectors~\cite{detr,sparsercnn,adamixer}, the decoder of DEQDet only has two different decoder layers: an initialization layer and an implicit refinement layer.
As shown in~\cref{fig:eq_view}, after the coarse object predictions are generated by the initialization layer, they are passed through this implicit refinement layer with infinite steps of iterations.
The object query refinement is represented as \textit{infinite-level fix-point solving process} of an \textit{implicit layer}, which could be solved by any black-box solver and enjoy the analytical backward pass independent of forward pass trajectories.
This fixed-point modeling perspective would share several advantages: (i) it would greatly scale up the modeling capacity of query-based detectors and is more flexible to deal with the large-variations of object instances. (ii) it would not rely on the traditional BPTT algorithms of RNN without storing the hidden states, which can save the memory overhead and make the training more efficient. (iii) it is a general modeling framework that could be applied to different query-based object detectors for detection performance improvement.

When training DEQDet, we find it is important to inject the refinement awareness~(\ie, the ability to perceive the operation that is performed iteratively) into its model parameter update. However, the commonly used methods for computing the gradients of implicit layers, such as Jacobian-free backpropagation~(JFB)~\cite{fung2022jfb}, lack the refinement awareness.
To tackle this problem, we propose to solve the two-step unrolled equilibrium equation with two new designs:
i)~Refinement-Aware Gradient~(RAG).
Through analysis, we find the refinement awareness corresponds to the high-order terms of Neumann-series expansion of inverse Jacobian term~\cite{geng2021training}, as they simulate the gradients propagated along the reverse of the solving path.
Therefore, we impose the \textit{refinement gradient term}, \ie, second-order terms of Neumann-series expansion, into the gradients. 
ii)~Refinement-Aware Perturbation~(RAP).
To further enhance the refinement awareness, we perturb the \textit{fixed-point} solving path by injecting noise.
Compared with merely adding Gaussian noise to each decoder layer,
our perturbation can be transmitted continuously with the iterations.
Thus, the deep supervision of query-based detectors on the perturbed solving path can encourage the refinement layer to be aware of its refinement nature.

We verify the effectiveness of our DEQDet framework on MS-COCO validation dataset by following the common practice. Our experiments demonstrate that DEQDet converges faster, consumes less memory, and achieves better results than the baseline counterpart (AdaMixer). In particular, our DEQDet with ResNet50 backbone and 300 queries achieves the $49.5$ \textit{mAP} and $33.0$ \textit{AP$_s$} on the MS COCO benchmark under $2\times$ training scheme (24 epochs). We also perform in-depth ablation study on the design of DEQDet and verify its scaling performance with stronger backbones such as Swin-S. Our \textbf{contributions} are as follows:
\begin{itemize}
    \item   We introduce RNN view over the query-based object detector and propose to model it as a fixed-point of an implicit layer with infinite depth.
    \item We propose Refinement Aware Gradient for DEQ model applied in high level semantic understanding task with sparsity nature like object detection.
    \item We propose Refinement Aware Perturbation to simulate the real noise of \textit{fixed-point} iterations in order to further improve refinement awareness of DEQ model. 
    \item Our experiments demonstrate the DEQDet achieves the state-of-the-art performance under a fair setting on the MS-COCO dataset.
\end{itemize}

% !TEX root = ../main.tex

\section{Related Work}
\label{sec:related}
\paragraph{Refinement in object detection.} 
The framework of two-stage object detectors~\cite{fasterrcnn,fastrcnn} can be deemed as a refinement-based detection paradigm. In these detectors, an \textit{initialization layer}, \eg, RPN~\cite{fasterrcnn}, is first adopted to generate some proposals which provide rough locations of objects. Then, a \textit{refinement layer}~(\ie the detection head formed by the region-wise feature extractor and the convolutional network) is employed to achieve precise localization and categorization for the object proposals.
Multi-stage object detectors like Cascade R-CNN~\cite{cascadercnn} introduce the \textit{multi-step refinement} into object detection. Cascade R-CNN adopts a series of detection heads to gradually refine the bounding boxes of objects to enable the high quality detection. 
Query-based detectors~\cite{detr,deformabledetr,sparsercnn,adamixer,dabdetr,smcadetr,interactor} are proposed to perform object detection through a set of \textit{learnable object queries}. These detectors are also formed by cascade decoder layers. In each layer, the image features are extracted by feature samplers and integrated into the input queries to generate the \textit{intermediate representations}. These representations can not only serve as the input of the next layer for further refinement, but can also be decoded into the class labels and the bounding box coordinates~\cite{detr}~(or coordinate offsets)~\cite{sparsercnn,adamixer} in current layer.
Despite the great success of the query-based object detectors, these detectors are unable to guarantee the input and output intermediate representations of each layer to lay in the same latent space, because the decoder layers are not \textit{weight-tied}~\cite{deq}. This design is less parameter-efficient. In addition, it is hard to determine whether they has achieved the convergence of refinement.  
%%%%%%%%%%%%%%%%%%%%%%%%%%%%%%%%%%%%%%%%%%%%%%%%%%
DiffusionDet~\cite{diffusiondet} borrows the denoising training technique from diffusion models~\cite{ddpm} into the refinement process of object detection, 
However, their refinement process is naively on the superficial space~(formed by bounding boxes) instead of the latent representations. 
%%%%%%%%%%%%%%%%%%%%%%%%%%%%%%%%%%%%%%%%%%%%%%%

\paragraph{Deep implicit neural network.}
Implict modeling has been explored by deep learning community by decades. Different from conventional neural networks that stack neural operators explicitly, implicit network defines its output by the solution of dynamic system. RBP~\cite{liao2018reviving, pineda1987generalization} trains the recurrent system implicitly by differentiation techniques. Neural ODE~\cite{chen2018neural} employs black-box ODE solvers to model recursive residual block implicitly. Deep Equilibrium Model (DEQ) ~\cite{deq,bai2020multiscale,fung2022jfb, bai2022deep, wang2020implicit} defines an implicit layer of solving fixed point equation to corresponding to infinite depth. Our DEQDet aims to leverage this modeling power of implicit DEQ to the specific challenging object detection task and propose customized optimization techniques to improve its training effectiveness and efficiency for object detection.

% !TEX root = ../main.tex
\section{Methodology}
\label{sec:methodology}

In general, our DEQDet can be applied to any query-based object detector. In current version, our DEQDet is mainly based on AdaMixer~\cite{adamixer}, a state-of-the-art query-based object detector which employs dynamic mixing in decoder design.
We first present a brief introduction of AdaMixer. Then, in order to introduce our DEQDet, we present a RNN perspective to reveal the refinement nature of decoder layer. After that we formulate the detection decoder as a \textit{fix-point} iteration process and propose our DEQDet. Finally we propose the training strategy of DEQDet. 

\subsection{AdaMixer Revisited}

\begin{figure}[t]
    % \vskip 0.2in
    \centering
    \includegraphics[width=.9\linewidth]{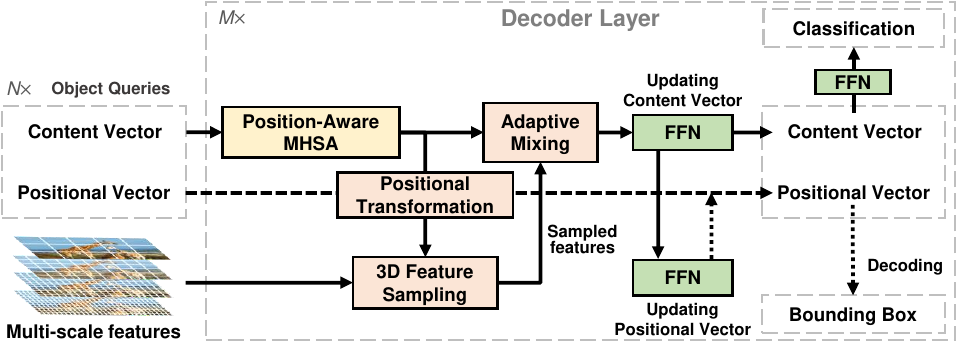}
    \caption{The detailed structure of AdaMixer~\cite{adamixer} decoder layer. The object query is decoupled into a content vector and a positional vector. The decoder operates on these two types of vectors and refine them through a dynamic 3D feature sampling module and an adaptive mixing module.}
    \label{fig:adamixer}
\end{figure}

Given an image input $I \in R^{3\times H \times W}$, object detectors are required to output object bounding box and its corresponding class category. The query-based object detector use a backbone with or without a neck encoder to extract multi-scale image features $\mathbf{x} = \{x^1, x^2, ..., x^l\}$, where $x^i \in R^{D\times {H^i} \times {W^i}}$ is the $i$-th level feature map and $l$ is the number of feature levels. 
Then, the features $\mathbf{x}$ with some learnable object queries are sequentially passed through a decoder containing $T$ independent decoder layers $\{f_1, f_2, ..., f_T\}$. The specific decoding process can be formulated as follows:
\begin{equation}
    \label{eq:ffn}
    \mathbf{y}_{t} = f_t(\mathbf{x}, \mathbf{y}_{t-1} | \theta_t),
\end{equation}
where $\mathbf{y}_{t}$ denotes the queries (or termed as the latent variables) at step $t$ outputted by layer $f_t$, and $\theta_t$ is the corresponding parameters of layer $f_t$. The main differences among current detectors are the definition of their object query $\mathbf{y}$ and the design of decoder layer $f$. Next, we will give a brief introduction to the AdaMixer design, and the detailed structure of AdaMixer decoder layer is illustrated in \cref{fig:adamixer}.

\paragraph{Object query of AdaMixer.} 
In the AdaMixer object detector, an object query is decomposed into two parts: a content query vector and a positional query vector:
\begin{equation}
    \mathbf{y}_t = (\mathbf{p}_t, \mathbf{q}_t),
\end{equation}
where $\mathbf{q}_t \in \mathbb{R}^D$ is the content vector of $\mathbf{y}_t$, and $\mathbf{p}_t$ is the corresponding positional vectors.
The content vector is expected to encode the appearance of object instances.
The positional vectors represent the coordinates of an individual bounding box.
Specifically, the positional vector in~\cite{adamixer} is parameterized as $(x,y,z,r) \in \mathbb{R}^4$, and its relation with the bounding box is as follows:
\begin{equation}
\begin{aligned}
x = x_{box} ~,& ~~~~ y = y_{box} ~,\\
z = \log_2 (\sqrt{w h}) ~,& ~~~~ r = \log_2 (\frac{h}{w}) ~,
\end{aligned}
\end{equation}
where $(x_{box}, y_{box})$ denotes the coordinates of the center point of the bounding box, and $w,h$ indicate the width and height of this box.

\paragraph{Decoder layer of AdaMixer.} 
As shown in~\cref{fig:ffn_view}, the object queries are sequentially passed through the decoder layers to refine features and boxes.
Each decoder layer of AdaMixer is typically composed of a multi-head self-attention module, a multi-head dynamic interaction module and some feed-forward networks~(FFN), as illustrated in~\cref{fig:adamixer}.
 
The object queries are first fed into the multi-head self-attention module, where the pairwise interaction is performed among queries.
Then, the outputs, \ie, the updated content vectors, are fed into the dynamic interaction module (3D feature sampling and adaptive mixing).
In this module, a set of image features are first sampled from the extracted multi-scale features according to the object queries, and then the adaptive mixing are performed on these features.
The processed features are then added into the content vectors.

Subsequently, each updated vector is fed into FFNs to predict the relative scaling and offsets to the positional vector~(for generating a new bounding box) and the classification scores.
Finally, the updated positional vectors~(bounding boxes) and content vectors are sent into the next decoder layer.

\subsection{DEQDet}
\label{sec:decoder}

After introducing the AdaMixer detector from a FNN view, we are ready to propose our DEQDet to improve it from both aspects of parameter efficiency and modeling capacity. We first reformulate AdaMixer from a RNN perspective to improve its parameter efficiency, and then further extend the modeling capacity of RNNDet to infinite refinement with deep equilibrium decoder.

\paragraph{{Decoder from FFN to RNN.}} 
As depicted in \cref{fig:ffn_view} and stated in \cref{eq:ffn}, the original query-based object detector works in a feed-forward way, where different decoder layers do not share weights and the resulting query vectors have different feature spaces. We argue this mechanism leads to large numbers of parameters and might be prone to overfitting. Instead, we observe that each decoder layer shares the same architecture and performs the progressive refinement of query vectors.
In this view, RNN might be a more parameter-efficient solution by incorporating weight-tying mechanism among different refinement layers, as shown in~\cref{fig:rnn_view}. In practice, this weight-tying strategy turns out not only the parameter is efficient, but also the detection performance can be improved.

Specifically, RNN iteratively processes the inputs with the same transformation and parameters. Formally, given a data sequence $[\mathbf{x}_0, \mathbf{x}_1, ..., \mathbf{x}_T]$ over time length (refinement step) $T$, the RNN decoder layers (except for initialization layer) share the same basic mathematical representation:
\begin{equation}
    \label{eq:rnn}
    \mathbf{y}_{t} = f(\mathbf{x}_{t}, \mathbf{y}_{t-1} | \theta),
\end{equation}
where
$\theta$ is the parameters of the RNN function $f$, $\mathbf{y}_t$ is the latent variable produced by the function $f$ at the time step $t$, and $\mathbf{y}_{t-1}$ is the preceding latent variable at the time step $t-1$.
Actually, the RNNDet processes a special data sequence, where every data item $\mathbf{x}_i$ is set to be the same multi-scale features $\mathbf{x}$ and $\mathbf{y}_t$ represents the object queries.
RNN is typically optimized through the BPTT~\cite{bptt}. The gradient flow of BPTT is illustrated in~\cref{fig:bptt}.

\paragraph{{Decoder from RNN to DEQ.}} Since RNN performs identical transformation on the inputs, the number of the iterations in~\cref{eq:rnn} can be easily extended to the infinity if all the $\mathbf{x}_{t}$ share the same value.
Furthermore, according to~\cite{deq}, when the sufficient stability condition is satisfied, the outputs of the weight-sharing layers of a general neural network tend to converge to a stable state as the model depth increases to infinity. In other words, when $t \to \infty$, the refinement layer would bring ``diminishing return'' and the network reaches an equilibrium:
\begin{equation}
    \lim_{t \to \infty} \mathbf{y}_t = \lim_{t \to\infty}f(\mathbf{x}, \mathbf{y}_{t} | \theta) \triangleq \mathbf{y}^*,
\end{equation}
where $\mathbf{y}^*$ indicates the fixed point (or called the equilibrium representation, the infinite feature representation).
We can directly solve this fixed point as a root-finding problem~\cite{deq}:
\begin{equation}
    \label{eq:fix_point}
    \mathbf{y}^* = f(\mathbf{x}, \mathbf{y}^*|\theta) .
\end{equation}
With this formulation, we can perform analytical backward pass in a constant memory consumption without tracing through the forward root-finding process.

\paragraph{The deep equilibrium decoder.}
To scale up detector into infinite-level refinement, we build our DEQDet based on the fixed point of an implicit layer. The overview of DEQDet architecture is presented in~\cref{fig:eq_view}.
In our framework, there are only two types of layers in our decoder: an \textit{initialization layer} and a \textit{refinement layer}.
The initialization layer first takes object queries as the input, and generates the image-related content vectors with image features and the coarse bounding box predictions:
\begin{equation}
    \mathbf{y}_{0} = g(\mathbf{x}, \mathbf{y}_{0^-} | \eta) ,
    \label{eq:initial_func}
\end{equation}
where $\mathbf{x}$ denotes the multi-scale image features extracted from the backbone (\eg, like a conventional neural network~\cite{resnet,resnext} or a vision transformer~\cite{swin,van}), the function $g$ refers to the initialization layer with $\eta$ as parameters, $\mathbf{y}_{0^-}$ denotes the initial object queries, and $\mathbf{y}_{0}$ denotes the object queries after initialization layer.
The refinement layer is an implicit layer which models the infinite refinement, as define in~\cref{eq:fix_point}, and its output is the fixed-point of this implicit layer.
To solve the value of $\mathbf{y}^*$, we can resort to naive solver or quasi-Newton methods (\eg, Anderson mixing~\cite{anderson_solver}), and set $\mathbf{y}_{0}$ as the initial value in these methods.

\subsection{Training of DEQDet}

We first introduce the gradients of an implicit layer and present a tractable approximation method which is widely used in previous works. 
Then, we propose our \textit{refinement aware gradient} (RAG) and \textit{refinement aware perturbation} (RAP) for the effective training of our DEQDet.

\paragraph{Gradients of an implicit layer.}
To differentiate through the implicit layer defined by \cref{eq:fix_point}, the gradient of $\theta$ and $\mathbf{x}$ under $\mathbf{y}^*$ can be derived from Implicit Function Theorem (IFT) as follows:
\begin{equation}
    \frac{\partial \mathbf{y}^* }{ \partial (\cdot)} = \left[ I - \frac{\partial f(\mathbf{x}, \mathbf{y}^*|\theta)} {\partial \mathbf{y}^*}\right]^{-1}\frac{\partial f(\mathbf{x}, \mathbf{y}^*|\theta)}{\partial {(\cdot)}},
    \label{eq:ift_grad}
\end{equation}
where the variables in $(\cdot)$ can be $\mathbf{x}$ or $\theta$, and the inverse-jacobian term $\left[ I - \frac{\partial f(\mathbf{x}, \mathbf{y}^*|\theta) }{ \partial \mathbf{y}^*}\right]^{-1}$ is the most intriguing part in gradient solving.
The original DEQ model integrates this term with VJP automation differential mechanism.
VJP transforms the gradient solving to another linear \textit{fixed-point system}, and thus it can also be solved via a \textit{fixed-point solver} off the shelf~\cite{deq}.
However, the fixed-point iteration for the gradient solving requires huge computational consumption, thereby prohibiting the application for real scenarios~\cite{geng2021training}.

\begin{figure}[t]
    \centering
    \subfloat[RNN Gradient flow\label{fig:bptt}]{
    \includegraphics[width=.99\linewidth]{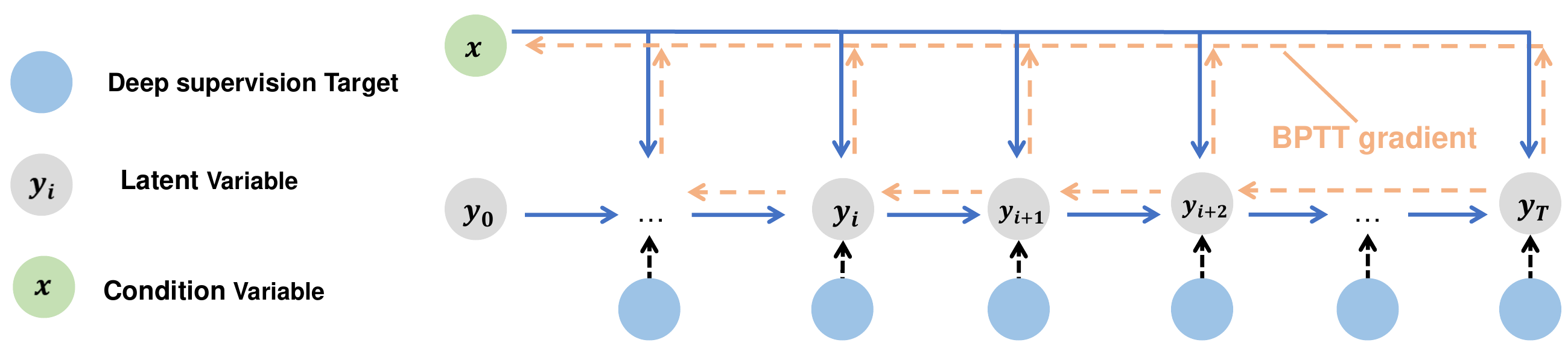}
    } \\
    \subfloat[DEQ Gradient flow\label{fig:RAG}]{
    \includegraphics[width=.99\linewidth]{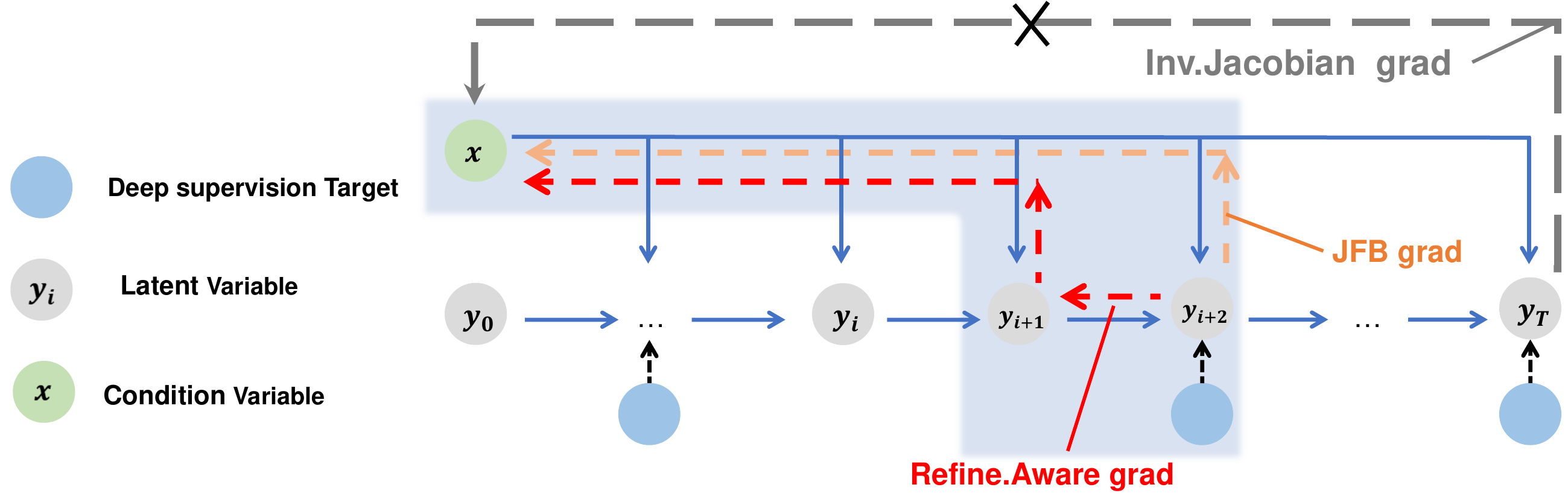}
    }
    \caption{\textbf{Gradient flow of DEQ model and RNN model}. The blue lines indicate forward flow. The dashed lines indicate the gradient flow. We use sparse deep supervision to train our DEQDet.}
    \vspace{-1em}
\end{figure}

\paragraph{Approximation of the inverse-Jacobian term.}
Following the recent works on backward gradient solving of implicit model ~\cite{fung2022jfb,bai2022deep}, we turn to estimate the inverse jacobian term because the resource consumption of the estimation method is relatively more acceptable. Specifically, Jacobian Free Backpropagation~(JFB)~\cite{fung2022jfb,bai2022deep} approximates the gradient formula~\cref{eq:JFB} by simply replacing the inverse jacobian term $\left[I - \frac{\partial f(x, y^*) }{ \partial y^*}\right]^{-1}$ in \cref{eq:ift_grad} with identity matrix $I$:
\begin{equation}
    \frac{\partial \mathbf{y}^* }{\partial (\cdot)} = I \cdot \frac{\partial f(\mathbf{x}, \mathbf{y}^*|\theta) }{\partial {(\cdot)}}.
    \label{eq:JFB}
\end{equation}
Although JFB avoids the overhead of inverse gradient calculations and achieves good results in some tasks~\cite{bai2022deep}, in fact such a simple estimation ignores the \textit{refinement} property of the function. The JFB gradient does not capture the relationship between input query $\mathbf{y}$ and updated query $f(\mathbf{x},\mathbf{y} | \theta)$. Therefore, in certain tasks that require high-level semantic understanding or have the sparsity, such as object detection, adopting JFB is not satisfactory.

\paragraph{Refinement-aware gradient and deep supervision.}
\label{sec:refinement_gradient}
To capture the \textit{refinement} nature of the decoder layer, we extend the \cref{eq:fix_point} to a two-step unrolled equilibrium equation:
\begin{equation}
     \label{eq:fix_point_unroll2}
     \mathbf{y}^* = f(\mathbf{x}, f(\mathbf{x}, \mathbf{y}^* | \theta) | \theta).
\end{equation} 
Based on \cref{eq:fix_point_unroll2}, we propose our refinement aware gradient (detailed derivation in Appendix):%\cref{app:rag}):
\begin{equation}
    \label{eq:rag}
    \frac{\partial \mathbf{y}^* }{ \partial (\cdot)} \approx \left[I + \frac{\partial f(\mathbf{x}, \mathbf{y}^* | \theta) }{\partial {\mathbf{y}^*}}\right] \frac{\partial f(\mathbf{x}, \mathbf{y}^* | \theta)}{ \partial {(\cdot)}}.
\end{equation}
Note that \cref{eq:rag} is also equivalent to 2-step Neumann-series-based Phantom Gradient~\cite{geng2021training}.
Exactly, $\frac{\partial f(\mathbf{x}, \mathbf{y}^* | \theta)}{ \partial \mathbf{y}^*}$ is the refinement gradient term.
Thus, in the Neumann-series of the inverse Jacobian term, its term $\small{\sum_{i=1}^k [\frac{\partial f(\mathbf{x}, \mathbf{y}^* | \theta) }{\partial \mathbf{y}^*}]^i}$ controls the refinement awareness.
We illustrate our refinement aware gradient with others in \cref{fig:RAG}.

As DEQ-Flow~\cite{bai2022deep} suggests, employing sparse deep supervision on the fixed point solving path can improve the model performance. 
From the perspective of optimal transport, the refinement layer tries its best to transfer the input queries $\mathbf{y_t}$ to our desired queries $\mathbf{\hat y}$ whose decoding boxes are identical to the true distribution in the given image. Thus at each refinement step $t$, the refinement layer will deliver the most closet value $\mathbf{y}_{t+1}$ to the desired queries $\mathbf{\hat y}$. This suggests the smaller t, the greater difficulty, Therefore, we choose to construct supervision positions set $\Omega$ in following way:
\begin{equation}
\label{eq:multiple_pos}
    \Omega_{\text{multiple}} = \{1, C, 2C, ... , m*C, T\} ,
\end{equation}
where $m,C$ are constant numbers.

\paragraph{Refinement-aware perturbation.}
\label{sec:RAP}
To further enhance the refinement awareness and improve the robustness of DEQDet, we introduce refinement-aware perturbation.

\textbf{\textit{A general way of adding Gaussian noise.}}
A simple noise-based perturbation is achieved by adding random noise to the latent variable $\mathbf{y}$, allowing the networks to recover from the corrupt result:
% A general way of this process is:
\begin{equation}
\label{eq:simple_noise}
    \hat{\mathbf{y}}_{n} = \mathbf{y}_{n} + \epsilon , ~~ \epsilon \sim \mathcal{N}( \mathbf{0},\sigma^2 I )
\end{equation}
where $\mathcal{N}(\cdot, \cdot)$ denotes a Gaussian distribution, $\sigma$ denotes the noise scale, and $\epsilon$ is the random variable sampled from this distribution.
However, directly adding this random noise is hard to simulate the real noise in fixed point solving process.
To tackle this, we introduce a new refinement-aware perturbation approach. We propose to use the refinement Jacobian matrix $\frac{\partial \mathbf{y}_n }{ \partial \mathbf{y}_{n-1}}$ to project a random noise to the latent space: % of $\mathbf{y}_n$:
\begin{equation}
\label{eq:one_step_noise}
    \hat{\mathbf{y}}_{n} = \mathbf{y}_n + \frac{\partial \mathbf{y}_n }{ \partial \mathbf{y}_{n-1}} \cdot \epsilon , ~~ \epsilon \sim \mathcal{N}( \mathbf{0},\sigma^2 I ),
\end{equation}
This approach can also be extended to a multi-step refinement-aware perturbation:
\begin{equation}
\label{eq:multi_step_noise}
    \hat{\mathbf{y}}_n = \mathbf{y}_n + \sum_{m=0}^{n-1} \mathbbm{1}_{ m \in \Psi} \cdot \frac{\partial \mathbf{y}_n }{\partial \mathbf{y}_{m}} \cdot \epsilon , ~~ \epsilon \sim \mathcal{N}( \mathbf{0},\sigma^2 I ),
\end{equation}
where $\Psi$ is the perturbation position set, indicating the indices of the solving path added with noise. This set is generated by probabilistic sampling, like random masking.

\textbf{\textit{Adding noise to object detector.}}
As for the specific implementation of adding noise to our detector, 
we treat the content vector $\mathbf q$ and positional vector $\mathbf p$ in different ways, as their physical meaning is not identical.
For the positional vector, we first decode it to the corner-format bounding box~(top-left point and bottom-right point), and then we add noise to these two points.
Note that this noise-adding operation may cause the flip between these two corner points.
Last, we transform the noise boxes to noise positional vectors.
As for the content vector, we construct Gaussian noise with variance $\Vert \mathbf{q} \Vert_2^2$, linearly mixing the noise and content vector with the perturbation size $\sigma_q$:
\begin{align}
\hat{\mathbf{q}} = (1-\sigma_{q})\mathbf{q} + \sigma_{q} \epsilon , ~~ \epsilon \sim \mathcal{N}( \mathbf{0}, \Vert \mathbf{q} \Vert_2^2 I ).
\end{align}
To impose the refinement Jacobian matrix on the noise term, in practice, we choose to directly feed the noisy latent variables into the refinement layer.
Then, the gradients provided by the noise term is equivalent to have the refinement Jacobian matrix as the multiplier. The detailed demonstration and the noise perturbation algorithm can be found in Appendix.

% !TEX root = ../main.tex

\begin{table}[t]
    \centering
    \small
    \resizebox{0.48\textwidth}{!}{
    \begin{tabular}{x{75}x{18}x{18}x{18}x{18}x{18}x{18}x{18}}
        \toprule
        Detectors  & AP  & AP$_{50}$ & AP$_{75}$ & AP$_s$ & AP$_m$& AP$_l$  \\
        \midrule
         FCOS~\cite{fcos}  & 38.7 & 57.4 & 41.8 & 22.9 & 42.5 & 50.1 \\
         Cascade R-CNN~\cite{cascadercnn}  & 40.4 & 58.9 & 44.1 & 22.8 & 43.7 & 54.0 \\
         GFocalV2~\cite{gfl_v2}  & 41.1 & 58.8 & 44.9 & 23.5 & 44.9 & 53.3 \\
         BorderDet~\cite{borderdet} & 41.4 & 59.4 & 44.5 & 23.6 & 45.1 & 54.6 \\
         Dynamic Head~\cite{dynamichead_det} & 42.6 & 60.1 & {46.4} & {26.1} & {46.8} & 56.0 \\
         DETR~\cite{detr} & 20.0 & 36.2 & 19.3 & 6.0  & 20.5 & 32.2 \\
        Deform-DETR~\cite{deformabledetr}  & 35.1 & 53.6 & 37.7 & 18.2 & 38.5 & 48.7 \\
        Sparse R-CNN~\cite{sparsercnn} & 37.9 & 56.0 & 40.5 & 20.7 & 40.0 & 53.5 \\
        \midrule
        AdaMixer$_{T=6}$\cite{adamixer} & {42.7} & {61.5} & {45.9} & {24.7} & {45.4} & {59.2}\\
        AdaMixer$^\dagger_{T=6}$\cite{adamixer} & {42.7} & {61.5} & {46.1} & {24.9} & {45.5} & {59.3}\\
        RNNDet$_{T=6}$ & {43.4} & {62.0} & {46.5} & {26.3} & {46.1} & {58.8}\\
        % RNNDet$_{T=8}$ &  {44.2} & {62.9} & {47.9} & {26.5} & {47.2} & {60.1}\\
        RNNDet$_{T=12}$ & {44.2} & {63.1} & {47.7} & {26.1} & {47.0} & {60.0}\\
        %RNNDet$_{T=16}$ & {43.8} & {62.7} & {47.2} & {26.9} & {46.7} & {59.1}\\
        % RNNDet$_{T=20}$ & {43.7} & {62.5} & {47.0} & {26.5} & {46.4} & {59.6}\\
        \midrule
        \textbf{DEQDet\space} & \textbf{45.3} & \textbf{64.0} & \textbf{48.9} & \textbf{27.7} & \textbf{47.9} & \textbf{61.5} \\
       \textbf{DEQDet$^\dagger$}  & \textbf{46.0} & \textbf{64.8} & \textbf{49.6} & \textbf{{27.5}} & \textbf{{49.0}} & \textbf{61.4} \\
        
       \bottomrule
        \end{tabular}
        }
    \caption{\textbf{Comparison with other detectors under classic $\mathbf{1}\times$ training scheme with 100 queries}. RNNDet and DEQDet consist of Initialization layer and Refinement layer and trained with deep supervision. $^\dagger$ means all layers with 64 sampling points instead of 32 sampling points }
    \vskip -0.1in
    \label{tab:1x}
\end{table}

\section{Experiments}
\label{sec:experiment}
We conduct experiments on the MS-COCO 2017 dataset \cite{coco}. The training batch size is set to 16. We employ AdamW optimizer~\cite{adamw} to update the parameters with weight decay $0.01$ for backbone and weight decay $0.1$ for decoder, The loss consists of focal loss~\cite{focalloss} with loss weight  $\lambda_\text{focal}=5$, L1 loss with loss weight $\lambda_\text{L1}=5$ and GIoU loss~\cite{giouloss} with loss weight $\lambda_\text{giou}=2$ . The matching cost for label assignment is aligned with loss. By default, we use the \textit{fixed-point} iteration steps $T_\text{train}=20$ for training and $T_\text{infer}=25$ for inference as there is just little performance gain in further increasing $T_\text{infer}$ . We place the detailed refinement steps experiment in Appendix. The base learning rate during training is $2.5\times10^{-5}$, and the lr multiplier for decoder is 4, we report the \textit{mAP} performance on COCO \textit{minival} set~\cite{coco}.

\begin{table*}[t]
    \centering
    \small
    \vskip 0.1in
    \resizebox{0.85\textwidth}{!}{
    \begin{tabular}{x{50}x{12}x{12}x{12}x{12}x{12}x{12}x{12}x{25}x{12}x{25}x{25}x{32}}
        \toprule
        Detectors & NF & AP  & AP$_{50}$ & AP$_{75}$ & AP$_s$ & AP$_m$ & AP$_l$ & Params & FPS & Mem$_\text{Train}$ & Mem$_\text{Infer}$ & TrainTime \\
        \midrule
        AdaMixer\cite{adamixer} & 6 & {42.7} & {61.5} & {45.9} & {24.7} & {45.4} & {59.2} &134M& 13.5 & 5961M & 872M & $\sim14.0$h\\
        AdaMixer$^\dagger$\cite{adamixer}& 6 & {42.7} & {61.5} & {46.1} & {24.9} & {45.5} & {59.3} &160M& 13.5 & 6803M & 972M & $\sim15.5$h  \\
        \midrule
        % SQR-AdaMixer$_{N=6}$ & {44.2} & {63.2} & {47.8} & {25.7} & {47.4} & {60.2}\\
        % SQR-AdaMixer$_{N=7}$ & {45.3} & {63.8} & {49.0} & {26.8} & {48.1} & {62.2}\\
        \multirow{5}*{RNNDet}& 6& {43.4} & {62.0} & {46.5} & {26.3} & {46.1} & {58.8} &61M& 13.6 & 4517M & 588M & $\sim12.0$h \\
        & 8& {44.2} & {62.9} & {47.9} & {26.5} & {47.2} & {60.1} &61M& 12.0 &4784M& 588M& $\sim13.5$h\\
        &12& {44.2} & {63.1} & {47.7} & {26.1} & {47.0} & {60.0} &61M& 9.2  & 5567M& 588M& $\sim17.5$h \\
        & 16&{43.8} & {62.7} & {47.2} & {26.9} & {46.7} & {59.1} &61M& 7.7 & 6195M& 588M& $\sim22.0$h \\
        & 20&{43.7} & {62.5} & {47.0} & {26.5} & {46.4} & {59.6} &61M& 5.9 & 6818M& 588M& $\sim24.0$h \\
        \midrule
       \multirow{4}*{\textbf{DEQDet\space}}&
       6& 44.3 & 63.0 & 47.6 & 26.2 & 47.1& 60.5& 61M& 13.6 & \multirow{4}*{4827M} &  \multirow{4}*{588M} & \multirow{4}*{$\sim25.5$h}  \\
       & 8 & 44.9 & 63.6 & 48.2 & 27.0& 47.5 & 61.1  & 61M & 12.0 &  \\
       & 16 & 45.2 & 63.9 & 48.8 & 27.5 & 47.9 & 61.3 & 61M & 7.7 &  \\
       & 25 & \textbf{45.3} & \textbf{64.0} & \textbf{48.9}& \textbf{27.7} & \textbf{47.9} & \textbf{61.5} &61M&5.1 &  \\
       
       \multirow{2}*{\textbf{DEQDet$^\dagger$}} & 
       6 & 45.7 & 64.3 & 49.3 & 27.5 & 48.7 & 61.9 & 69M& 13.0& \multirow{2}*{4997M} & \multirow{2}*{622M} & \multirow{2}*{$\sim29.0$h} \\
       & 25 & \textbf{46.0} & \textbf{64.8} & \textbf{49.6}& \textbf{27.5} & \textbf{49.0} &\textbf{61.4}& 69M & 4.8&   \\
        \bottomrule
        \end{tabular}
        }
    \caption{\textbf{classic $\mathbf{1}\times$ training results} with 100 queries. RNNDet and DEQDet consist of Initialization layer and Refinement layer and trained with deep supervision. $^\dagger$ means all layers with 64 sampling points instead of 32 sampling points }
    \vskip -0.1in
    \label{tab:1x_details}
\end{table*}

\subsection{Classic $1\times$ Training Results}
\begin{figure}[t]
    \centering
    \includegraphics[width=0.98\linewidth]{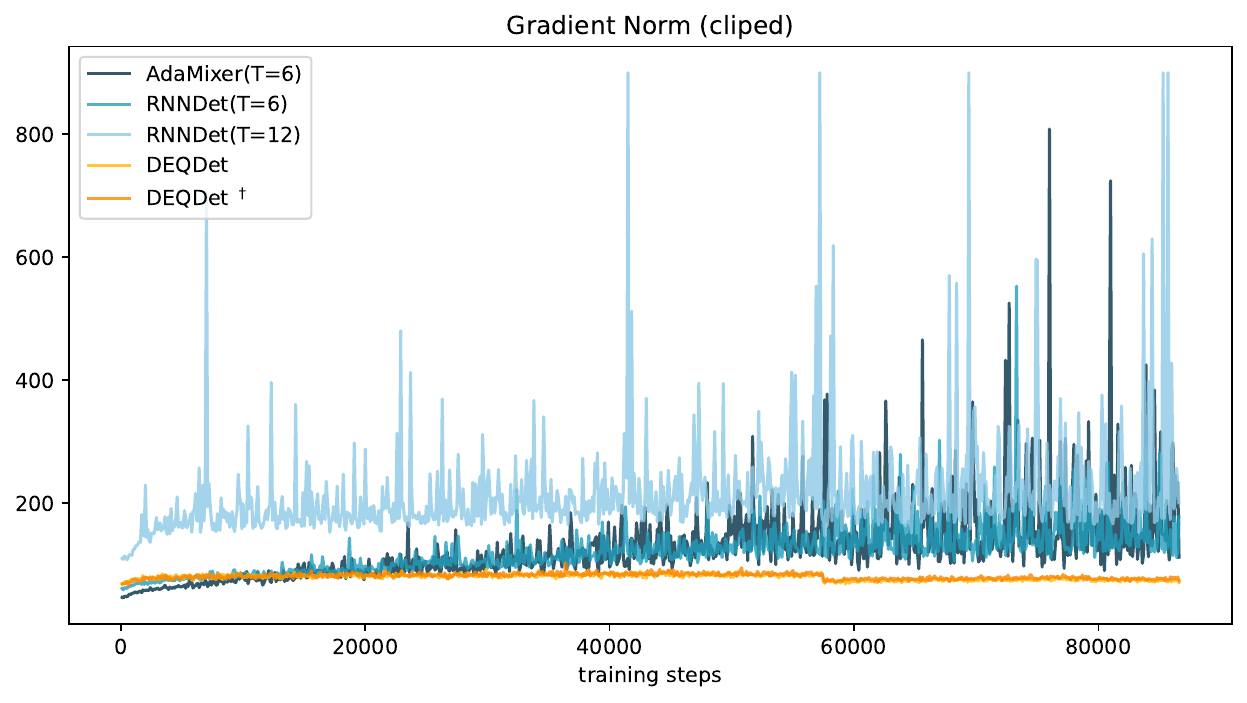}
    \caption{\textbf{Gradient Norm of Detectors}, To obtain the gradient norm , we first flatten all gradients as a single vector, then calculate the l2-norm of this gradient vector. }
    \label{fig:grad_norm}
    \vskip -0.1in
\end{figure}

We first report the performance of DEQDet by adopting the classic $1 \times$ training scheme. The classic $1\times$ training scheme contains 12 training epochs with training images of shorter side resized to 800 and only with random flip data augmentation. In this study, the object query number is set to 100. We present the detailed results of $1\times$ training results of RNNDet and DEQDet in \cref{tab:1x_details} and compare with other detectors in \cref{tab:1x}. First, we compare the results between AdaMixer and RNNDet. With the same number of refinement (NF=6), RNNDet obtains the better performance than AdaMixer (43.4 vs. 42.7) with less than half parameters and similar inference speed. This superior performance verifies the effectiveness of weight-tying strategy. Second, we increase the refinement number in RNNDet from 6 to 20, and obtain the best performance of 44.2 at NF= 8 or 12. However, we can clearly observe that as the number of refinement layers in RNNDet further increases, the performance degrades partially due to RNN optimization difficulty. We visualize the gradient norm in~\cref{fig:grad_norm} and the RNN norm is not stable. Then, we present the result of our DEQDet and see the gradient
norm of DEQDet is very consistent and stable. When NF in the fixed-point solving process is set to 6, our DEQDet achieves better performance (44.3) with less parameters and smaller memory consumption than AdaMixer. When we further increase the NF in DEQDet to 25, it obtains the best performance of 45.3 under 32 sampling points and 46.0 under 64 sampling points. Finally, we notice that the inference time of DEQDet is comparable to the other counterparts, but its training time is relatively larger due to the forward fixed-point solving process.

We also compare our DEQDet with other detectors under this limited training epochs and data augmentations in~\cref{tab:1x}. The results demonstrate that our DEQDet achieves significant improvement over previous detectors under this limited training budget. This result show that our DEQDet is training-efficient and provides a highly competitive baseline for future object detector design. 

\begin{figure}[t]
    \vspace{-1mm}   
    \centering
    \includegraphics[width=0.98\linewidth]{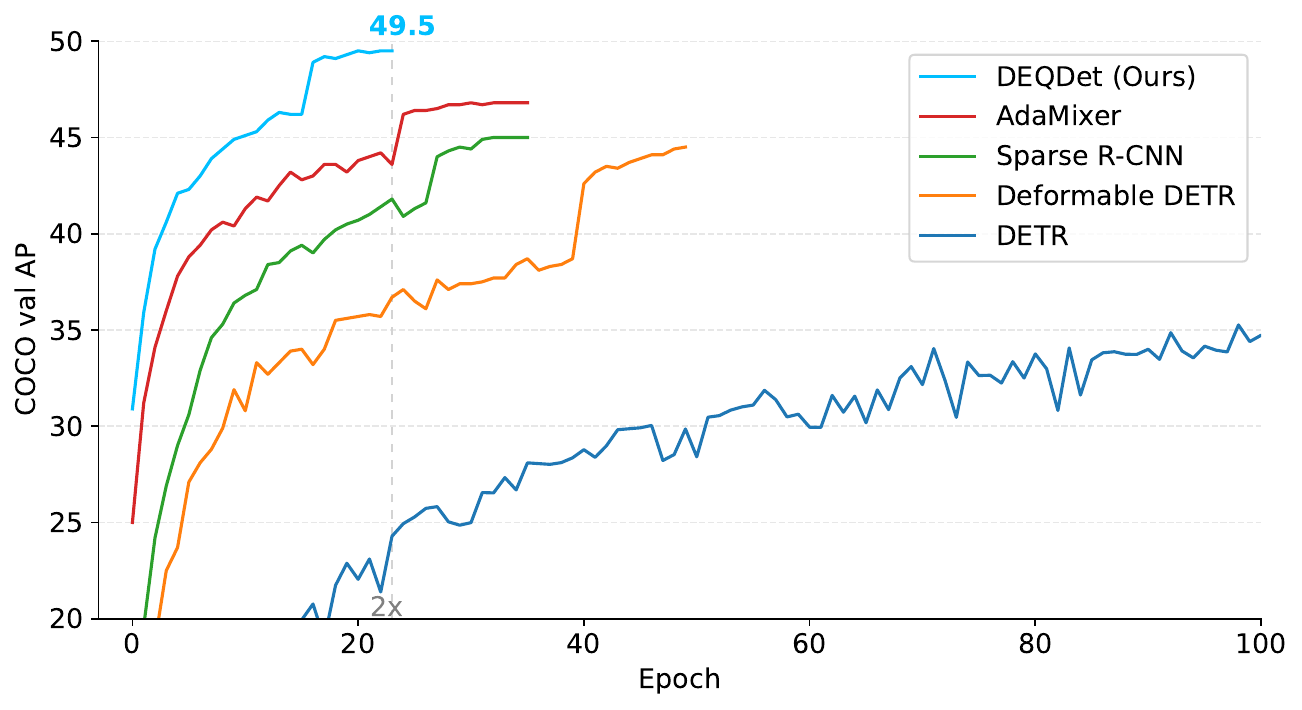}
    \caption{\textbf{Trainng Convergence Curves} of DEQDet and AdaMixer~\cite{adamixer}, Sparse-RCNN~\cite{sparsercnn}, DETR~\cite{detr}, Deformable DETR~\cite{deformabledetr}. The number of object queries is 300 and backbone is ResNet50.}
    \vspace{-2mm}
    \label{fig:training_curve}
\end{figure}

\subsection{Comparison with the state of the art}
\begin{table*}[t]
    \centering
    \small
    \setlength{\tabcolsep}{4pt}
    \vskip 0.1in
    \resizebox{0.95\textwidth}{!}{
    \begin{tabular}{l|c|c|x{24}|x{30}|x{12}x{12}x{12}x{12}x{12}x{12}}
        \toprule
        Detector & Backbone & {Encoder/FPN} & { Epochs} & { Params} & AP  & AP$_{50}$ & AP$_{75}$ & AP$_{s}$ & AP$_m$ & AP$_l$  \\
        \midrule
        DETR~\cite{detr} & ResNet-50-DC5 & TransformerEnc & 500 & 41M & 43.3 & 63.1 & 45.9 & 22.5 & 47.3 & 61.1  \\
        SMCA~\cite{smcadetr} & ResNet-50 & TransformerEnc & 50 & 40M & 43.7 & 63.6 & 47.2 & 24.2 & 47.0 & 60.4 \\
        Deformable DETR~\cite{deformabledetr} & ResNet-50 & DeformTransEnc & 50 & 40M & 43.8 & 62.6 & 47.7 & 26.4 & 47.1 & 58.0  \\
        Anchor DETR~\cite{anchordetr} & ResNet-50-DC5 & DecoupTransEnc & 50 & 35M & 44.2 & 64.7 & 47.5 & 24.7 & 48.2 & 60.6 \\
        Efficient DETR~\cite{efficientdetr} & ResNet-50 & DeformTransEnc & 36 & 35M & 45.1 & 63.1 & 49.1 & 28.3 & 48.4 & 59.0 \\
        Conditional DETR~\cite{conditionaldetr} & ResNet-50-DC5 & TransformerEnc & 108 & 44M & 45.1 & 65.4 & 48.5 & 25.3 & 49.0 & {62.2}\\
        Sparse R-CNN~\cite{sparsercnn} & ResNet-50 & FPN & 36 & 110M  & 45.0 & 63.4 & 48.2 & 26.9 & 47.2 & 59.5  \\
        REGO~\cite{glimpsedetr} & ResNet-50 & DeformTransEnc & 50 & 54M & 47.6 & 66.8 & 51.6 & 29.6 & 50.6 & 62.3 \\
        DAB-D-DETR~\cite{dabdetr} & ResNet-50 & DeformTransEnc & 50 & 48M & 46.8 & 66.0 & 50.4 & 29.1 & 49.8 & 62.3 \\
        DN-DAB-D-DETR~\cite{dndetr} & ResNet-50 & DeformTransEnc & 12 & 48M & 43.4 & 61.9 & 47.2 & 24.8 & 46.8 & 59.4 \\ 
        DN-DAB-D-DETR~\cite{dndetr} & ResNet-50 & DeformTransEnc & 50 & 48M & 48.6 & {67.4} & 52.7 & 31.0 & 52.0 & 63.7 \\ 
        AdaMixer~\cite{adamixer} & ResNet-50 & - & 12 & 139M & 44.1 & 63.1 & 47.8 & 29.5 & 47.0 & 58.8  \\
        AdaMixer~\cite{adamixer} & ResNet-50 & - & 24 & 139M & 46.7 & 65.9 & 50.5 & 29.7 & 49.7 & 61.5  \\ 
        AdaMixer~\cite{adamixer} & ResNet-50 & - & 36 & 139M & {47.0} & {66.0} & {51.1} & {30.1} & {50.2} & {61.8}  \\
        RNNDet$_{T=8}$ & ResNet-50 & -  & 36 & 65M  & 48.1 & 66.7 & 52.3 & 31.2 & 51.1 & 62.5 \\
        RNNDet$^\dagger_{T=8}$ & ResNet-50 & -  & 36 & 69M& 48.4 & 67.1 & 52.7 & 31.8 & 51.4 & 63.4 \\
        DEQDet & ResNet-50 & - & 12 & 65M  & {46.6}  & {65.3}  &  {50.6} & {30.5}  & {49.4}  &  {61.2}  \\
        DEQDet & ResNet-50 & - & 24 & 65M  & \textbf{48.6} & \textbf{67.6} & \textbf{53.0} & \textbf{31.6} & \textbf{51.8} & \textbf{62.9}  \\
        DEQDet$^\dagger$ & ResNet-50 & - & 24 & 69M  & \textbf{49.5} & \textbf{68.1} & \textbf{53.9} & \textbf{33.0} & \textbf{52.0} & \textbf{63.3}  \\
        \midrule
        DETR~\cite{detr} & ResNet-101-DC5 & TransformerEnc & 500 & 60M & 44.9 & 64.7 & 47.7 & 23.7 & 49.5 & 62.3 \\
        SMCA~\cite{smcadetr} & ResNet-101 & TransformerEnc & 50 & 58M & 44.4 & 65.2 & 48.0 & 24.3 & 48.5 & 61.0 \\
        Efficient DETR~\cite{efficientdetr} & ResNet-101 & DeformTransEnc  & 36 & 54M & 45.7 & 64.1 & 49.5 & 28.2 & 49.1 & 60.2 \\
        Conditional DETR~\cite{conditionaldetr} & ResNet-101-DC5 & TransformerEnc & 108 & 63M & 45.9 & 66.8 & 49.5 & 27.2 & 50.3 & 63.3 \\
        Sparse R-CNN~\cite{sparsercnn} & ResNet-101 & FPN & 36 & 125M & 46.4 & 64.6 & 49.5 & 28.3 & 48.3 & 61.6 \\
        REGO~\cite{glimpsedetr} & ResNet-101 & DeformTransEnc & 50 & 73M & 48.5 & 67.0 & 52.4 & 29.5 & 52.0 & 64.4 \\
        AdaMixer~\cite{adamixer} & ResNet-101 & -  & 36 & 158M & {48.0} & {67.0} & {52.4} & {30.0} & {51.2} & {63.7} \\
        DEQDet & ResNet-101 & -  & 24 & 84M  & \textbf{49.5} & \textbf{68.2} & \textbf{53.8} & \textbf{33.6} & \textbf{52.8} & \textbf{64.3} \\ 
        DEQDet$^\dagger$ & ResNet-101 & -  & 24 & 88M  & \textbf{50.1} & \textbf{68.9} & \textbf{54.5} & \textbf{34.3} & \textbf{53.3} & \textbf{65.1} \\
        \midrule
        AdaMixer~\cite{adamixer} & Swin-S & -  & 36 & 164M & {51.3} & 71.2 & {55.7} & {34.2} & {54.6} & {67.3} \\
        DEQDet & Swin-S & -  & 24 & 90M  & \textbf{52.7} & \textbf{72.3} & \textbf{57.6} & \textbf{36.6} & \textbf{55.9} & \textbf{68.4} \\ 
        \bottomrule
        \end{tabular}
        }
    \caption{Comparison with other detectors on COCO \textit{minival }set. The number of queries defaults to 300 in our DEQDet. $^\dagger$ means refinement layer with 64 sampling points.} 
    \label{tab:sota}
    \vskip -0.1in
\end{table*}
After reporting the results under limited training budget, we will scale up our DEQDet with more object queries, longer training epochs, and stronger backbones. First, we visualize the training convergence curves of R50 backbone and 300 queries in \cref{fig:training_curve}. DEQDet is designed to explore the  potential power of refinement layer as much as possible. Compared with our counterpart AdaMixer~\cite{adamixer}, DEQDet convergences faster and achieves higher \textit{mAP}. We conjecture the gradient norm in DEQDet is consistent and stable as illustrated in~\cref{fig:grad_norm},  which leads to faster convergence. 

We compare the results of our DEQDet with other detectors in \cref{tab:sota}. In \cref{tab:sota}, we allocate 300 queries in DEQDet  Our DEQDet shows fast convergence and thus we only scale the training epochs to 24, which is smaller than previous methods. We observe that under the backbone of ResNet50, our DEQDet$^\dagger$ achieves $49.5$ \textit{mAP} under $2\times$ training scheme. outperforming its baseline Adamixer by $2.5$ \textit{mAP}. Especially in small object detection metrics, DEQDet$^\dagger$ achieves $33.0$ \textit{AP$_s$}. We also scale the backbone of our DEQDet to ResNet-101 and Swin-S. Our DEQDet can outperform the AdaMixer by 2.1 \textit{mAP} for ResNet101 and 1.4 \textit{mAP} for Swin-S. This shows our DEQDet generalizes well to large backbones.

\subsection{Ablation studies}
In this ablation study, we use 100 queries and ResNet50 as the backbone for DEQDet. The training epoch is 12. 
\begin{table*}[t]
\centering
\newcommand\tr{\rule{0pt}{10pt}}
\newcommand\br{\rule[-3.3pt]{0pt}{3.3pt}}
%#################################################
% Initialization layer
%#################################################
\subfloat[
\textbf{Init Layer}. A large Init layer benifits performance.
\label{tab:init_stage}
]{
\setlength{\tabcolsep}{1.5pt}
\centering
\begin{minipage}{0.2\linewidth}{\begin{center}
\tablestyle{2pt}{1.05}
\begin{tabular}{x{36}x{18}x{18}x{18}}
\shline
Init layer sampl.points & AP & AP$_{50}$ & AP$_{75}$ \\
\shline
/ & 45.2 & 63.8 & 48.8 \\
32 & 45.3 & 64.0 & 48.9 \\
{64} & {{45.5}} & {{64.4}} & {{49.1}} \\
\shline
\end{tabular}
\end{center}}
\end{minipage}
}
\hspace{1.5em}
%#################################################
% Initialization layer
%#################################################
\subfloat[
\textbf{Init Layer supervised} with extra refinement aware gradient.
\label{tab:init_RAG}
]{
\centering
\begin{minipage}{0.2\linewidth}{\begin{center}
\tablestyle{2pt}{1.05}
\setlength{\tabcolsep}{.45pt}
\begin{tabular}{x{36}x{18}x{18}x{18}}
\shline
extra super. layers & AP & AP$_{50}$ & AP$_{75}$ \\
\shline
0 & 44.7 & 63.4 & 48.2 \\
1 & 45.1 & 63.8 & 48.8 \\
{2} & {{45.5}} & {{64.4}} & {{49.1}} \\
\shline
\end{tabular}
\end{center}}
\end{minipage}
}
\hspace{1.5em}
%#################################################
% refinement gradient
%#################################################
\subfloat[
\textbf{refinement aware gradient} with different $k$.
\label{tab:refinement_gradient}
]{
\centering
\begin{minipage}{0.2\linewidth}{\begin{center}
\tablestyle{2pt}{1.05}
\begin{tabular}{x{36}x{18}x{18}x{18}}
\shline
RAG step.k & AP & AP$_{50}$ & AP$_{75}$ \\
\shline
1    & 41.9 & 60.7 & 45.0 \\
{2} & {45.5} & {64.4} & {49.1} \\
3 & 45.5 & 64.2 & 49.4 \\
4 & 45.7 & 64.2 & 49.9 \\
\shline
\end{tabular}
\end{center}}
\end{minipage}
}
\hspace{1.5em}
%#################################################
% deep supervision position
%#################################################
\subfloat[
\label{tab:deep_superpos} 
\textbf{deep supervision position set $\Omega$.} % $m=4$ with $C=3$ achieves the best performance.
]{
\centering
\begin{minipage}{0.2\linewidth}{\begin{center}
\tablestyle{1.3pt}{1.05}
\begin{tabular}{x{18}x{18}x{18}x{18}x{18}}
\shline
 m & C & AP & AP$_{50}$ & AP$_{75}$ \\
 \shline
% 4 & \multirow{2}*{*}  & 45.3 & 64.1 & 49.0\\
% 5 & {~} & 45.1 & 63.8 & 48.9 \\
% \shline
% 6 & random & 45.4 & 64.0 & 49.1 & 28.9 & 48.3 & 59.8\\
{4} & {3} & {45.5} & {64.4} & {49.1} \\
4 & 4 & 45.4 & 63.9 & 49.2 \\
% 3 & 4 & 45.2 & 64.0 & 48.9\\
3 & 3 & 45.1 & 64.0 & 48.9\\
5 & 3 & 45.2 & 63.8 & 48.8 \\ 
\shline
\end{tabular}
\end{center}}
\end{minipage}
}
\vspace{0.8em}
%#################################################
% perturbation step
%#################################################
\subfloat[
\textbf{perturbation step} multiple step perturbation works best.
]{
\label{tab:perturb_step}
\begin{minipage}{0.2\linewidth}{
\begin{center}
\tablestyle{1.3pt}{1.05}
\begin{tabular}{x{36}x{18}x{18}x{18}}
\shline
\tr{perturbation step.}\br & AP & AP$_{50}$ & AP$_{75}$ \\
\shline
zero-step & 44.9 & 63.8 & 48.6 \\
one-step & 45.1 & 63.9 & 48.6 \\
{multi-step} & {{45.5}} & {{64.4}} & {{49.1}}  \\
\shline
\end{tabular}
\end{center}
}
\end{minipage}
}
\hspace{1.5em}
% %#################################################
% % single noise 
% %#################################################
\subfloat[
\textbf{single perturbation noise.} % Both postional noise and content noise improve the performance.
\label{tab:perturb_noise}
]{
\centering
\begin{minipage}[t]{0.2\linewidth}
{
\begin{center}
\tablestyle{2pt}{1.05}
\begin{tabular}{x{12}x{12}x{18}x{18}x{18}}
\shline
$\sigma_q$ & $\sigma_p$ & AP & AP$_{50}$ & AP$_{75}$ \\
\shline
/ & / & 44.3 & 63.1 & 47.8 \\
/ & 25 & 45.4 & 64.2 & 49.0 \\
/ & 50 & 45.5 & 64.5 & 49.1\\
0.1 & / & 45.0 & 63.6 & 48.5\\
0.2 & / & 45.2 & 64.3 & 48.6\\
\shline
\end{tabular}
\end{center}
}
\end{minipage}
}
\hspace{1.5em}
%#################################################
% combined noise 
%#################################################
\subfloat[
\textbf{perturbation noise combinations.}  
% The noise combinations of best single noise do not yield the best result. Excessive noise hurts the detector.
\label{tab:perturb_noise_comb}
]{
\centering
\begin{minipage}{0.2\linewidth}
{\begin{center}
\tablestyle{2pt}{1.05}
\begin{tabular}{x{18}x{18}x{18}x{18}x{18}}
\shline
$\sigma_q$ & $\sigma_p$ & AP & AP$_{50}$ & AP$_{75}$ \\
\shline
{0.1} & {25} & {45.5 }& {64.4} & {49.1}\\
0.1 & 50 & 45.5 & 64.4 & 49.1\\
0.2 & 25 & 45.2 & 63.9 & 49.0\\
0.2 & 50 & 45.2 & 63.9 & 49.2\\
\shline
\end{tabular}
\end{center}}
\end{minipage}
}
\hspace{1.5em}
%#################################################
% T_train
%#################################################
\subfloat[
\textbf{\textit{fixed-point}~iteration steps} during training.
\label{tab:forward_num}
]{
\centering
\begin{minipage}{0.2\linewidth}{\begin{center}
\tablestyle{2pt}{1.05}
\begin{tabular}{x{36}x{18}x{18}x{18}}
\shline
iteration steps & AP & AP$_{50}$ & AP$_{75}$ \\
\shline
 15   & 45.1 & 63.8 & 48.7 \\
{20}  & {{45.5}} & {{64.4}} & {{49.1}} \\
  25 & 45.3 & 64.0 & 48.9 \\
\shline
\end{tabular}
\end{center}}
\end{minipage}
}
\vskip -0.05in
\caption{\textbf{Ablation Studies} on our DEQDet design with ResNet-50 as the backbone and 100 object queries under the 12 training epochs on the MS-COCO \textit{minival} set. }
\vspace{-3mm}
\end{table*}

\paragraph{Initialization layer.}
As introduced in \cref{sec:decoder}, we employ an initialization layer to convert the image content agnostic queries to image content-related queries. We investigate different initialization layer setting in \cref{tab:init_stage}, including 1). no initialization layer, 2). an initialization layer with 32 sampling points 3). an initialization layer with 64 sampling points. As initialization layer with enough sampling points can obtain relatively rich semantic information from input features, it will releive the learning difficulty of subsequent layers. Another question is whether to apply extra supervision to initialization layer during training. We summarize $h=0,1,2$ in \cref{tab:init_RAG}, where $h$ represents the number of extra layers placed on top of initialization layer for supervision. When $h=0$, there is no connection between initialization layer and refinement layer, so the training is unstable. 

\paragraph{Refinement-aware gradient.} As discussed in \cref{sec:refinement_gradient}, due to the high-level semantic understanding property and sparse nature of object detection, ignoring the refinement aware gradient will lead to sub-optimal results. We verify our conjecture through experiments in the \cref{tab:refinement_gradient}. The refinement aware gradient step $k$ refers to the expansion steps of neumann-series. When $k=1$, RAG degenerates into JFB~\cite{fung2022jfb}, which fails to consider the refinement property of \textit{fixed-point} iteration. $k=2$ is the standard RAG baseline, which achieves $45.5$ mAp. RAG outperforms JFB substantially in object detection task. Although the $k=3$ setting retains the locality property and enjoys more refinement awareness, it offers little improvement. So, we set $k=2$.

\paragraph{Deep supervision position $\Omega$.} 
\label{sec:exp_deep_supervision}
We experiment with position set in \cref{eq:multiple_pos} under different settings. We use the default RAG step $k=2$.  We experimentally find using $\Omega_{\text{multiple}}$ with $m=4, C=3$ achieves the best result. 

\paragraph{Refinement-aware perturbation.} 
We compare zero-step simple noise and perturbation noise in \cref{tab:perturb_step}.  As introduced in \cref{sec:RAP}, one-step perturbation noise naively projects the simple noise by refinement layer to latent space. One-step noise improves the simple noise by $0.2$ mAP, which means adding the noise associated with fixed-point solving is better than simple Gaussian noise. The best results are achieved with multi-step noise, as it takes full advantage of the fixed-point solution path.

\paragraph{Perturbation noise.}
In \cref{tab:perturb_noise}, we experiment with different noise scales in terms of position noise perturbation and content perturbation and their combinations. From \cref{tab:perturb_noise}, we find both content perturbation and position perturbation significantly improve the performance of DEQDet. The improvement effect of position perturbation is more obvious than that of content perturbation. As the noise scale increases, the performance can be further improved. However, \cref{tab:perturb_noise_comb}, when we combine the best content noise $\sigma_q = 0.2$ and the best best position noise $\sigma_p = 50$, DEQDet yields only $45.2$ \textit{mAP}. The Performance degradation of perturbation combination shows excessive noise perturbation hurts DEQDet performance. An reasonable noise scale $\sigma_q = 0.1$ and $\sigma_p = 25$ achieves the best \textit{mAP} ($45.5$).

\paragraph{Refinement iteration steps.} 
We also experiment with different \textit{fixed-point} iteration steps $T_\text{train}$ during training of DEQDet. \cref{tab:forward_num} shows $T_\text{train}=20$ achieves the highest mAP performance($45.5$). Comparing other steps, we can conclude that more refinement steps during training can enhance the performance of detectors. But $T_\text{train}=25$ does not exceed $T_\text{train}=20$, which may be due to limited deep supervision positions for those redundant refinements.

% !TEX root = ../main.tex

\section{Conclusion}
In this paper, we have proposed the deep equilibrium detector (DEQDet), a query-based object detector with infinite refinement steps.
We equivalently model the refinement process as a \textit{fixed-point} solving problem of a implicit layer.
As for the training of DEQDet, we find that its simple estimation on inverse-jacobian term lacks refinement awareness, resulting in a negative impact on the high-level semantic understanding of the object detector.
Therefore, to inject the refinement awareness into the detector during training, we propose the refinement-aware gradient~(RAG) and the refinement-aware perturbation~(RAP). Our experiments show DEQDet converges faster, consumes less memory, and achieves better performance than the counterparts on the MS-COCO dataset.
We hope our DEQDet becomes a strong baseline and could inspire future work to consider deep equilibrium modeling in other computer vision tasks.
\label{sec:conclusion}

\paragraph {\bf Acknowledgements.} {\small This work is supported by National Key R$\&$D Program of China (No. 2022ZD0160900), National Natural Science Foundation of China (No. 62076119, No. 61921006).}

{\small
\bibliographystyle{ieee_fullname}
\bibliography{main}

\begin{thebibliography}{10}\itemsep=-1pt

\bibitem{anderson_solver}
Donald~G Anderson.
\newblock Iterative procedures for nonlinear integral equations.
\newblock {\em Journal of the ACM (JACM)}, 12(4):547--560, 1965.

\bibitem{bai2022deep}
Shaojie Bai, Zhengyang Geng, Yash Savani, and J~Zico Kolter.
\newblock Deep equilibrium optical flow estimation.
\newblock In {\em Proceedings of the IEEE/CVF Conference on Computer Vision and
  Pattern Recognition}, pages 620--630, 2022.

\bibitem{deq}
Shaojie Bai, J~Zico Kolter, and Vladlen Koltun.
\newblock Deep equilibrium models.
\newblock {\em Advances in Neural Information Processing Systems}, 32, 2019.

\bibitem{bai2020multiscale}
Shaojie Bai, Vladlen Koltun, and J~Zico Kolter.
\newblock Multiscale deep equilibrium models.
\newblock {\em Advances in Neural Information Processing Systems},
  33:5238--5250, 2020.

\bibitem{cascadercnn}
Zhaowei Cai and Nuno Vasconcelos.
\newblock Cascade r-cnn: high quality object detection and instance
  segmentation.
\newblock {\em IEEE transactions on pattern analysis and machine intelligence},
  43(5):1483--1498, 2019.

\bibitem{detr}
Nicolas Carion, Francisco Massa, Gabriel Synnaeve, Nicolas Usunier, Alexander
  Kirillov, and Sergey Zagoruyko.
\newblock End-to-end object detection with transformers.
\newblock In {\em Proceedings of the European conference on computer vision},
  pages 213--229, 2020.

\bibitem{chen2018neural}
Ricky~TQ Chen, Yulia Rubanova, Jesse Bettencourt, and David~K Duvenaud.
\newblock Neural ordinary differential equations.
\newblock {\em Advances in neural information processing systems}, 31, 2018.

\bibitem{diffusiondet}
Shoufa Chen, Peize Sun, Yibing Song, and Ping Luo.
\newblock Diffusiondet: Diffusion model for object detection.
\newblock {\em arXiv preprint arXiv:2211.09788}, 2022.

\bibitem{glimpsedetr}
Zhe Chen, Jing Zhang, and Dacheng Tao.
\newblock Recurrent glimpse-based decoder for detection with transformer.
\newblock In {\em Proceedings of the IEEE/CVF Conference on Computer Vision and
  Pattern Recognition}, pages 5260--5269, 2022.

\bibitem{dynamichead_det}
Xiyang Dai, Yinpeng Chen, Bin Xiao, Dongdong Chen, Mengchen Liu, Lu Yuan, and
  Lei Zhang.
\newblock Dynamic head: Unifying object detection heads with attentions.
\newblock In {\em Proceedings of the IEEE/CVF conference on computer vision and
  pattern recognition}, pages 7373--7382, 2021.

\bibitem{duan2019centernet}
Kaiwen Duan, Song Bai, Lingxi Xie, Honggang Qi, Qingming Huang, and Qi Tian.
\newblock Centernet: Keypoint triplets for object detection.
\newblock In {\em Proceedings of the IEEE/CVF international conference on
  computer vision}, pages 6569--6578, 2019.

\bibitem{fung2022jfb}
Samy~Wu Fung, Howard Heaton, Qiuwei Li, Daniel McKenzie, Stanley Osher, and
  Wotao Yin.
\newblock Jfb: Jacobian-free backpropagation for implicit networks.
\newblock In {\em Proceedings of the AAAI Conference on Artificial
  Intelligence}, 2022.

\bibitem{smcadetr}
Peng Gao, Minghang Zheng, Xiaogang Wang, Jifeng Dai, and Hongsheng Li.
\newblock Fast convergence of detr with spatially modulated co-attention.
\newblock In {\em Proceedings of the IEEE/CVF International Conference on
  Computer Vision}, pages 3621--3630, 2021.

\bibitem{adamixer}
Ziteng Gao, Limin Wang, Bing Han, and Sheng Guo.
\newblock Adamixer: A fast-converging query-based object detector.
\newblock In {\em Proceedings of the IEEE/CVF Conference on Computer Vision and
  Pattern Recognition}, pages 5364--5373, 2022.

\bibitem{geng2021training}
Zhengyang Geng, Xin-Yu Zhang, Shaojie Bai, Yisen Wang, and Zhouchen Lin.
\newblock On training implicit models.
\newblock {\em Advances in Neural Information Processing Systems},
  34:24247--24260, 2021.

\bibitem{fastrcnn}
Ross Girshick.
\newblock Fast r-cnn.
\newblock In {\em Proceedings of the IEEE international conference on computer
  vision}, pages 1440--1448, 2015.

\bibitem{van}
Meng-Hao Guo, Cheng-Ze Lu, Zheng-Ning Liu, Ming-Ming Cheng, and Shi-Min Hu.
\newblock Visual attention network.
\newblock {\em arXiv preprint arXiv:2202.09741}, 2022.

\bibitem{resnet}
Kaiming He, Xiangyu Zhang, Shaoqing Ren, and Jian Sun.
\newblock Deep residual learning for image recognition.
\newblock In {\em Proceedings of the IEEE conference on computer vision and
  pattern recognition}, pages 770--778, 2016.

\bibitem{ddpm}
Jonathan Ho, Ajay Jain, and Pieter Abbeel.
\newblock Denoising diffusion probabilistic models.
\newblock {\em Advances in Neural Information Processing Systems},
  33:6840--6851, 2020.

\bibitem{cornernet}
Hei Law and Jia Deng.
\newblock Cornernet: Detecting objects as paired keypoints.
\newblock In {\em Proceedings of the European conference on computer vision},
  pages 734--750, 2018.

\bibitem{dndetr}
Feng Li, Hao Zhang, Shilong Liu, Jian Guo, Lionel~M Ni, and Lei Zhang.
\newblock Dn-detr: Accelerate detr training by introducing query denoising.
\newblock In {\em Proceedings of the IEEE/CVF Conference on Computer Vision and
  Pattern Recognition}, pages 13619--13627, 2022.

\bibitem{gfl_v2}
Xiang Li, Wenhai Wang, Xiaolin Hu, Jun Li, Jinhui Tang, and Jian Yang.
\newblock Generalized focal loss v2: Learning reliable localization quality
  estimation for dense object detection.
\newblock In {\em Proceedings of the IEEE/CVF Conference on Computer Vision and
  Pattern Recognition}, pages 11632--11641, 2021.

\bibitem{liao2018reviving}
Renjie Liao, Yuwen Xiong, Ethan Fetaya, Lisa Zhang, KiJung Yoon, Xaq Pitkow,
  Raquel Urtasun, and Richard Zemel.
\newblock Reviving and improving recurrent back-propagation.
\newblock In {\em International Conference on Machine Learning}, pages
  3082--3091. PMLR, 2018.

\bibitem{focalloss}
Tsung-Yi Lin, Priya Goyal, Ross Girshick, Kaiming He, and Piotr Doll{\'a}r.
\newblock Focal loss for dense object detection.
\newblock In {\em Proceedings of the IEEE international conference on computer
  vision}, pages 2980--2988, 2017.

\bibitem{coco}
Tsung-Yi Lin, Michael Maire, Serge Belongie, James Hays, Pietro Perona, Deva
  Ramanan, Piotr Doll{\'a}r, and C~Lawrence Zitnick.
\newblock Microsoft coco: Common objects in context.
\newblock In {\em European conference on computer vision}, pages 740--755.
  Springer, 2014.

\bibitem{dabdetr}
Shilong Liu, Feng Li, Hao Zhang, Xiao Yang, Xianbiao Qi, Hang Su, Jun Zhu, and
  Lei Zhang.
\newblock Dab-detr: Dynamic anchor boxes are better queries for detr.
\newblock {\em arXiv preprint arXiv:2201.12329}, 2022.

\bibitem{ssd}
Wei Liu, Dragomir Anguelov, Dumitru Erhan, Christian Szegedy, Scott Reed,
  Cheng-Yang Fu, and Alexander~C Berg.
\newblock Ssd: Single shot multibox detector.
\newblock In {\em European conference on computer vision}, pages 21--37.
  Springer, 2016.

\bibitem{swin}
Ze Liu, Yutong Lin, Yue Cao, Han Hu, Yixuan Wei, Zheng Zhang, Stephen Lin, and
  Baining Guo.
\newblock Swin transformer: Hierarchical vision transformer using shifted
  windows.
\newblock In {\em Proceedings of the IEEE/CVF International Conference on
  Computer Vision}, pages 10012--10022, 2021.

\bibitem{adamw}
Ilya Loshchilov and Frank Hutter.
\newblock Decoupled weight decay regularization.
\newblock {\em arXiv preprint arXiv:1711.05101}, 2017.

\bibitem{conditionaldetr}
Depu Meng, Xiaokang Chen, Zejia Fan, Gang Zeng, Houqiang Li, Yuhui Yuan, Lei
  Sun, and Jingdong Wang.
\newblock Conditional detr for fast training convergence.
\newblock In {\em Proceedings of the IEEE/CVF International Conference on
  Computer Vision}, pages 3651--3660, 2021.

\bibitem{pineda1987generalization}
Fernando Pineda.
\newblock Generalization of back propagation to recurrent and higher order
  neural networks.
\newblock In {\em Neural information processing systems}, 1987.

\bibitem{borderdet}
Han Qiu, Yuchen Ma, Zeming Li, Songtao Liu, and Jian Sun.
\newblock Borderdet: Border feature for dense object detection.
\newblock In {\em European Conference on Computer Vision}, pages 549--564.
  Springer, 2020.

\bibitem{fasterrcnn}
Shaoqing Ren, Kaiming He, Ross Girshick, and Jian Sun.
\newblock Faster r-cnn: Towards real-time object detection with region proposal
  networks.
\newblock {\em Advances in neural information processing systems}, 28, 2015.

\bibitem{giouloss}
Hamid Rezatofighi, Nathan Tsoi, JunYoung Gwak, Amir Sadeghian, Ian Reid, and
  Silvio Savarese.
\newblock Generalized intersection over union: A metric and a loss for bounding
  box regression.
\newblock In {\em Proceedings of the IEEE/CVF conference on computer vision and
  pattern recognition}, pages 658--666, 2019.

\bibitem{ddim}
Jiaming Song, Chenlin Meng, and Stefano Ermon.
\newblock Denoising diffusion implicit models.
\newblock {\em arXiv preprint arXiv:2010.02502}, 2020.

\bibitem{sparsercnn}
Peize Sun, Rufeng Zhang, Yi Jiang, Tao Kong, Chenfeng Xu, Wei Zhan, Masayoshi
  Tomizuka, Lei Li, Zehuan Yuan, Changhu Wang, et~al.
\newblock Sparse r-cnn: End-to-end object detection with learnable proposals.
\newblock In {\em Proceedings of the IEEE/CVF conference on computer vision and
  pattern recognition}, pages 14454--14463, 2021.

\bibitem{teed2020raft}
Zachary Teed and Jia Deng.
\newblock Raft: Recurrent all-pairs field transforms for optical flow.
\newblock In {\em European conference on computer vision}, pages 402--419.
  Springer, 2020.

\bibitem{interactor}
Yao Teng, Haisong Liu, Sheng Guo, and Limin Wang.
\newblock {StageInteractor}: Query-based object detector with cross-stage
  interaction.
\newblock In {\em Proceedings of the IEEE/CVF Conference on Computer Vision and
  Pattern Recognition}, 2023.

\bibitem{fcos}
Zhi Tian, Chunhua Shen, Hao Chen, and Tong He.
\newblock Fcos: Fully convolutional one-stage object detection.
\newblock In {\em Proceedings of the IEEE/CVF international conference on
  computer vision}, pages 9627--9636, 2019.

\bibitem{wang2020implicit}
Tiancai Wang, Xiangyu Zhang, and Jian Sun.
\newblock Implicit feature pyramid network for object detection.
\newblock {\em arXiv preprint arXiv:2012.13563}, 2020.

\bibitem{anchordetr}
Yingming Wang, Xiangyu Zhang, Tong Yang, and Jian Sun.
\newblock Anchor detr: Query design for transformer-based detector.
\newblock In {\em Proceedings of the AAAI conference on artificial
  intelligence}, pages 2567--2575, 2022.

\bibitem{bptt}
Paul~J Werbos.
\newblock Backpropagation through time: what it does and how to do it.
\newblock {\em Proceedings of the IEEE}, 78(10):1550--1560, 1990.

\bibitem{resnext}
Saining Xie, Ross Girshick, Piotr Doll{\'a}r, Zhuowen Tu, and Kaiming He.
\newblock Aggregated residual transformations for deep neural networks.
\newblock In {\em Proceedings of the IEEE conference on computer vision and
  pattern recognition}, pages 1492--1500, 2017.

\bibitem{efficientdetr}
Zhuyu Yao, Jiangbo Ai, Boxun Li, and Chi Zhang.
\newblock Efficient detr: improving end-to-end object detector with dense
  prior.
\newblock {\em arXiv preprint arXiv:2104.01318}, 2021.

\bibitem{zhou2021centernet2}
Xingyi Zhou, Vladlen Koltun, and Philipp Kr{\"a}henb{\"u}hl.
\newblock Probabilistic two-stage detection.
\newblock {\em arXiv preprint arXiv:2103.07461}, 2021.

\bibitem{centernet}
Xingyi Zhou, Dequan Wang, and Philipp Kr{\"a}henb{\"u}hl.
\newblock Objects as points.
\newblock {\em arXiv preprint arXiv:1904.07850}, 2019.

\bibitem{deformabledetr}
Xizhou Zhu, Weijie Su, Lewei Lu, Bin Li, Xiaogang Wang, and Jifeng Dai.
\newblock Deformable detr: Deformable transformers for end-to-end object
  detection.
\newblock {\em arXiv preprint arXiv:2010.04159}, 2020.

\end{thebibliography}
}

\newpage
\appendix
\section{Notions and hyperparameters in DEQDet}
To avoid confusion by the notions in DEQDet, we summarize all symbols in~\cref{tab:hyperparameter}. For the convenience of others to reproduce the experiment in DEQDet, we also provides the training hyper-parameters.
\begin{table*}[htbp]
  %\small
  \renewcommand\arraystretch{1.2}
  \centering
  
  \vskip -0.5in
  \begin{tabular}{ll|c}
    \toprule
    Hyperparameters & Notation  & Value  \\
    \midrule
    The content vector & $\mathbf{q}$ & \\
    The positional vector &  $\mathbf{p}$ &  \\
    The condition variable / multi-scale features & $\mathbf{x}$ & \\
    The latent variable & $\mathbf{y}=(\mathbf{p}, \mathbf{q})$ & \\
    The parameters & $\theta, \eta$ & \\
    The refinement layer & $f(\mathbf{x}, \mathbf{y})$ & \\
    The initialization layer & $g(\mathbf{x}, \mathbf{y})$ & \\
    The deep supervision position set & $\Omega$ & [1,3,6,9,12,20]\\
    The number fixpoint iteration steps for training  & $T_\text{train}$ & 20\\
    The number fixpoint iteration steps for inference  & $T_\text{infer}$ & 25\\
    The perturbation probability & $v$ & 0.2 \\
    The perturbation size of content vector & $\sigma_q$ & 0.1\\
    The perturbation size of positional vector & $\sigma_p$ & 25\\
    The sampling points of initialization layer & & 64 \\
    The sampling points of refinement layer layer & & 32 \\
    The learning rate & & 0.000025 \\
    The learning rate decay & & *0.1 \\
    The learning rate decay epoch for $1\times$ training & & 8, 11 \\
    The learning rate decay epoch for $2\times$ training & & 16, 22 \\
    weight decay for backbone & & 0.01 \\
    weight decay for decoder & & 0.1 \\
    The loss weight for focal loss &  $\lambda_\text{focal}$ & 2 \\
    The loss weight for l1 loss &  $\lambda_\text{l1}$ & 5 \\
    The loss weight for giou loss &  $\lambda_\text{giou}$ & 2 \\
    \bottomrule
  \end{tabular}
  % \vspace{-1.0em}
  \caption{The hyper-parameters of DEQDet.}
  % \vskip -0.5in
  \label{tab:hyperparameter}
\end{table*}
\section{Noise projection for \textit{fixed-point}}
\label{app:noise_porjection}
To impose the refinement jacobian matrix on the noise term, in practice, we directly feed the noisy latent variables into the refinement layer. Then, the gradients provided
by the noise term is equivalent to have the refinement jacobian matrix as the multiplier. We conduct directly feeding noise to refinement layer instead of computing jacobian, due to
\begin{itemize}
    \item jacobian based projection is equivalent to the one step Taylor expansion of refinement layer.
    \item computing jacobian matrix spends more time
    \item actually, as detaching position vector in object detectors is a common practice, employing automatic differentiation library to solve jacobian matrix will delivery wrong results.
\end{itemize}

\begin{align}
    \hat{\mathbf{y}}_{n} &= f(x, {\mathbf{y}}_{n-1} + \epsilon) \\
     &\approx f(x, {\mathbf{y}}_{n-1}) + {\partial \mathbf{y}_n \over \partial \mathbf{y}_{n-1}} \cdot \epsilon \\
     &={\mathbf{y}}_{n} + {\partial \mathbf{y}_n \over \partial \mathbf{y}_{n-1}} \cdot \epsilon ~~ \epsilon \sim \mathcal{N}( \mathbf{0},\sigma^2 I )
\end{align}
\section{Refinement-aware gradient derivation}
\label{app:rag}
To handle Refinement awareness, we reformulate \textit{fixed-point} formula to a two-step unrolled \textit{fix-point} to take the query term into account:
\begin{equation}
     \label{eq:fix_point2}
     \mathbf{y}^* = f(\mathbf{x}, f(\mathbf{x}, \mathbf{y}^* | \theta) | \theta)
\end{equation} 
For simplicity, we define a new function $h$, which is the two-step unrolled refinement layer:
\begin{equation}
     \label{eq:fix_point2_h_func}
     h(\mathbf{x}, \mathbf{y}^* | \theta)=f(\mathbf{x}, f(\mathbf{x}, \mathbf{y}^* | \theta) | \theta)
\end{equation} 
Then, the IFT gradient of~\cref{eq:fix_point2} becomes:
\begin{equation}
    \label{eq:ift_grad2}
    {\partial \mathbf{y}^* \over \partial (\cdot)} = (I - {\partial h(\mathbf{x}, \mathbf{y}^* | \theta) \over \partial \mathbf{y}^*})^{-1}{\partial h(\mathbf{x}, \mathbf{y}^* | \theta) \over \partial (\cdot)} 
\end{equation}
We first replace the inverse jacobian term $(I - {\partial h(\mathbf{x}, \mathbf{y}^* | \theta) \over \partial \mathbf{y}^*})^{-1}$ with identity matrix $I$ as JFB,
and then we implicitly differentiate two sides of~\cref{eq:fix_point2_h_func}:
\begin{equation}
    \label{eq:ift_grad3}
    {\partial \mathbf{y}^* \over \partial (\cdot)} \approx {\partial h(\mathbf{x}, \mathbf{y}^* | \theta) \over \partial (\cdot)}  =[I + {\partial f(\mathbf{x}, \mathbf{y}^* | \theta) \over \partial {\mathbf{y}^*}}] {\partial f(\mathbf{x}, \mathbf{y}^* | \theta) \over \partial {(\cdot)}} 
\end{equation}
then we get our refinement aware gradient:
\begin{equation}
    {\partial \mathbf{y}^* \over \partial (\cdot)} \approx [I + {\partial f(\mathbf{x}, \mathbf{y}^* | \theta) \over \partial {\mathbf{y}^*}}] {\partial f(\mathbf{x}, \mathbf{y}^* | \theta) \over \partial {(\cdot)}} 
\end{equation}

\section{Connection between DEQ model and diffusion model}
There are some differences and connections between diffusion model and \textit{fixed-point} iterations based DEQ model, a basic \textit{fixed -point} form likes :
\begin{equation}
    y = f(x, y),
\end{equation}
\begin{equation}
    y_n = f(x, y_{n-1}),
\end{equation}
while diffusion~\cite{ddpm, ddim} can be derived from ode form (ddim~\cite{ddim}) :
\begin{equation}
    {dy \over dt} = g(x, y, t),
\end{equation}
when we make a finite integral for ode, we can get:
\begin{equation}
    y_n = y_{n-1} + \int_{t_{n-1}}^{t_n} g(x, y, t) dt,
\end{equation}
\begin{equation}
    y_n = y_{n-1} + g(x, y, t) \Delta t,
\end{equation}
so, it is clear that the function $f(x, y)$ in \textit{fixed-point} iteration can be any arbitrary form, instead \textbf{ode always keeps an identity branch or residual connection}. But actually we also use identity branch in our DEQDet decoder layer. The second difference is \textbf{ode is step aware} as $g(x,y,t)$ takes step $t$ as input, while \textit{fixed-point} iteration not. 
\section{Comparison with similar works}
\par Except our method mainly focuses on \textit{fixed-point} iteration, others devote to migrate diffusion diagram to object detection~\cite{diffusiondet}. The remaining major difference between our detector and DiffusionDet~\cite{diffusiondet} is that our decoder consists of only two layers, the first layers aims to get a good initial guess while the second layer progressively refines this initial result. Then the definition of a refinement step is also distinct. we regard running refinement layer once as a refinement step while their refinement step runs the entire decoder, which consists of 6 layers. 
\section{Extend DEQDet to other detector}
We also extend our DEQDet to sparse-RCNN ~\cite{sparsercnn}. we keep the training settings \eg training epochs, optimizer, learning rate scheduler consistent with the original sparse-rcnn, Our DEQDet improves the sparse-rcnn by $2.5$ on mAP and  $3.7$ on $\text{AP}_\text{small}$.
\begin{table}[htbp]
    \centering
    \small
    \renewcommand\arraystretch{1.0}
    \setlength{\tabcolsep}{3pt}
    \vskip -0.1in
    \begin{tabular}{x{80}x{20}x{15}x{15}x{15}x{15}x{15}x{15}}
        \toprule
        Detectors & Params & AP  & AP$_{50}$ & AP$_{75}$ & AP$_{s}$ & AP$_m$ & AP$_l$ \\
        \midrule
        Sparse R-CNN~\cite{sparsercnn} & 110M  & 45.0 & 63.4 & 48.2 & 26.9 & 47.2 & 59.5  \\
        +DEQDet (2x) & 53M  &
        {47.0} & {65.7} & {51.6} & {30.3} & {49.8} &{61.0} \\
        +DEQDet (3x) & 53M  &
        {47.5} & {66.5} & {52.4} & {30.6} & {50.1} &{61.5} \\
        \bottomrule
        \end{tabular}
    \caption{\textbf{{$\mathbf{3}\times$ training scheme} with 300 queries.} Extend DEQDet to other detectors \eg sparse-rcnn~\cite{sparsercnn}. }
    \vskip -0.1in
    \label{tab:extend}
\end{table}

\section{Refinement convergence}
\label{app:refinement_steps}
We evaluate our DEQDet with different refinment steps in \cref{tab:refine_steps100} and \cref{tab:refine_steps300}. We conduct experiments on DEQDet$^\dagger$ trained under $1\times$ scheme with 100 queries and $2\times$ scheme with 300 queries.  In \cref{tab:refine_steps300}, when DEQDet$^\dagger$ with 300 queries refines 5 steps, the number of valid decoder layers is as same as AdaMixer~\cite{adamixer}, but DEQDet achieves $49.0$ mAP, exceeds AdaMixer by $2.0$ mAP. As the refinement step increases, the performance will be further improved.

\begin{table}[htbp]
    \centering
    \small
    \renewcommand\arraystretch{1.0}
    \setlength{\tabcolsep}{3pt}
    \vskip -0.2in
    \begin{tabular}{x{36}x{20}x{20}x{20}x{20}x{20}x{20}x{20}x{20}}
        \toprule
        GFLOPS & steps & AP  & AP$_{50}$ & AP$_{75}$ & AP$_{s}$ & AP$_m$ & AP$_l$ \\
        \midrule\
         107.50 & 4  & 44.9 & 63.5 & 48.4 & 26.1 & 48.0 & 60.9 \\
         109.73 & 5  & 45.4 & 64.0 & 49.0 & 26.6 & 48.4 & 61.2 \\
         111.97 & 6  & 45.7 & 64.3 & 49.4 & 26.8 & 48.7 & 61.4 \\
         116.43 & 8  & 45.8 & 64.5 & 49.6 & 27.0 & 48.9 & 61.3 \\
         
         120.90 & 10  & 45.9 & 64.6 & 49.7 & 27.2 & 48.9 & 61.2 \\
         132.07 & 15  & 45.9 & 64.7 & 49.6 & 27.4& 49.0 & 61.2 \\
         143.24 & 20  & 46.0 & 64.7 & 49.6 & 27.4& 49.0 & 61.4 \\
         154.41 & 25  & 46.0 & 64.8 & 49.6 & 27.5 & 49.0 & 61.2 \\
         
         210.25 & 50  & 46.0 & 64.7 & 49.6 & 27.5 & 49.0 & 61.5 \\
         - & 200 & 46.0 & 64.8 & 49.7 & 27.6 & 49.1 & 61.5 \\
        \bottomrule
        \end{tabular}
    \caption{Refinement steps for DEQDet$^\dagger$ trained under $1\times$ scheme with ResNet50 backbone and 100 queries.}
    \vskip -0.2in
    \label{tab:refine_steps100}
\end{table}

\begin{table}[htbp]
    \centering
    \small
    \renewcommand\arraystretch{1.0}
    \setlength{\tabcolsep}{3pt}
    \vskip 0.1in
    \begin{tabular}{x{36}x{20}x{20}x{20}x{20}x{20}x{20}x{20}x{20}}
        \toprule
        GFLOPS & steps & AP  & AP$_{50}$ & AP$_{75}$ & AP$_{s}$ & AP$_m$ & AP$_l$ \\
        \midrule\
         129.87 & 4  & 48.7 & 67.3 & 52.8 & 32.0 & 51.7 & 63.0 \\
         136.57 & 5  & 49.0 & 67.7 & 53.3 & 32.3 & 51.9 & 63.0 \\
         143.27 & 6  & 49.1 & 67.8 & 53.5 & 32.5 & 52.0 & 63.2 \\
         156.67 & 8  & 49.3 & 68.0 & 53.7 & 32.9 & 52.0 & 63.2 \\
         
         170.07 & 10  & 49.5 & 68.2 & 53.9 & 33.1 & 52.1 & 63.4 \\
         203.58 & 15  & 49.5 & 68.3 & 53.9 & 33.2 & 52.1 & 63.3 \\
         237.08 & 20  & 49.5 & 68.3 & 54.0 & 33.2 & 52.1 & 63.4 \\
         270.59 & 25  & 49.5 & 68.3 & 54.0 & 33.2 & 52.1 & 63.3 \\
         
         438.11 & 50  & 49.6 & 68.3 & 54.0 & 33.3 & 52.2 & 63.1 \\
         - & 200 & 49.5 & 68.3 & 54.0 & 33.2 & 52.2 & 63.1 \\
        \bottomrule
        \end{tabular}
    \caption{Refinement steps for DEQDet$^\dagger$ trained under $2\times$ scheme with ResNet50 backbone and 300 queries.}
    \label{tab:refine_steps300}
\end{table}
\par We try to employ off-the-shelf \textit{fixed-point} solver \eg anderson solver to accelerate the \textit{fixed-point} solving, but the result is not what we expected. We think this is mainly due to the highly nonlinear property of the refinement layer and we should couple the solver with training instead of decoupling. We left this for our future work.
\begin{table}[htbp]
    \centering
    \small
    \renewcommand\arraystretch{1.0}
    \setlength{\tabcolsep}{3pt}
    \vskip 0.1in
    \begin{tabular}{x{36}x{20}x{20}x{20}x{20}x{20}x{20}x{20}x{20}}
        \toprule
        $m$ & steps & AP  & AP$_{50}$ & AP$_{75}$ & AP$_{s}$ & AP$_m$ & AP$_l$ \\
        \midrule
         % 2&4  &  48.5 & 67.2 & 52.7 & 31.8 & 51.5 & 62.7 \\
         2&5  &  48.9 & 67.5 & 53.1 & 32.2 & 51.8 & 63.1 \\
         2&10  & 49.2 & 68.0 & 53.6 & 32.9 & 52.0 & 63.1\\
         2&20  & 49.4 & 68.2 & 53.8 & 32.9 & 52.0 & 63.4 \\
         2&50  & 49.5 & 68.2 & 53.9 & 33.1 & 52.2 & 63.3 \\
         \midrule
         4&5   & 48.8 & 67.5 & 53.1 & 32.2 & 51.8 & 63.3 \\
         4&10  & 49.2 & 67.9 & 53.5 & 32.7 & 52.0 & 63.1 \\
         4&20  & 49.3 & 68.1 & 53.7 & 33.1 & 52.0 & 63.4 \\
         4&50  & 49.4 & 68.1 & 53.8 & 33.1 & 52.1 & 63.3 \\
        \bottomrule
        \end{tabular}
    \caption{Refinement steps of Anderson Solver for EQDet$^\dagger$ with ResNet50 backbone and 300 queries. $m$ is a hyperparameter in Anderson Solver.}
    % \label{tab:refine_steps}
\end{table}
\section{Detection Performance on COCO test set}
We also provide the detection performance of DEQDet models on COCO \textit{test-dev} set in \cref{tab:coco_test}. Different from the COCO \textit{minival} set, there is no publicly avaliable labels of \textit{test-dev}. 
\begin{table}[htbp]
\centering
\small
\renewcommand\arraystretch{1.0}
\setlength{\tabcolsep}{3pt}
\vskip -0.15in
\vspace{2mm}
\resizebox{0.47\textwidth}{!}{
\begin{tabular}{c c c c c c c c c c }
\hline
Detectors & Backbone & queries & AP & AP$_{50}$ & AP$_{75}$ & AP$_{s}$ & AP$_{m}$ & AP$_{l}$  \\
\hline
DEQDet (1x) & R50 & 100  & 45.4 & 64.5 & 49.0 & 26.0 & 47.7 & 59.3 \\
DEQDet$^\dagger$ (1x) & R50&100 & 46.5 & 65.5 & 50.4 & 27.2 & 48.9 & 60.2 \\
DEQDet$^\dagger$ (2x) & R50 &300& 49.8  & 68.5 & 54.4 & 31.2 & 52.1 & 62.5 \\
DEQDet$^\dagger$ (2x) & R101 &300& 50.6 & 69.4 & 55.1 & 31.3 & 53.3 & 64.2 \\
\hline
\end{tabular}
}
\vskip -0.1in
\caption{Detection Performance of DEQDet on COCO \textit{test-dev} set, $1\times$ means $1\times$ training scheme, including 12 epochs, while $2\times$ contains 24 epochs. }
\label{tab:coco_test}
\vskip -0.2in
\end{table}

\section{Limitations}
Although our DEQDet achieves comparable results with acceptable resource consumption, the training time consumption is still very large compared to other methods. As for inference time, it is acceptable to choose refinement steps adaptively according to resource constraints. There are also a lot of improvement space for training strategy. Also note that the refinement layer is not light and each refinement iteration is not really cheap. Future improvements can be made from light weight refinement layer design and reduction of refinement steps
\section{Training algorithm}
\label{app:algo}
\begin{algorithm}[htbp]
\small
\caption{ Noise Perturbation Code~\label{alg:noise}
}
% \algcomment{\fontsize{7.2pt}{0em}}
\definecolor{codeblue}{rgb}{0.25,0.5,0.5}
\definecolor{codegreen}{rgb}{0,0.6,0}
\definecolor{codekw}{RGB}{207,33,46}
\lstset{
  backgroundcolor=\color{white},
  basicstyle=\fontsize{7.5pt}{7.5pt}\ttfamily\selectfont,
  columns=fullflexible,
  breaklines=true,
  captionpos=b,
  commentstyle=\fontsize{7.5pt}{7.5pt}\color{codegreen},
  keywordstyle=\fontsize{7.5pt}{7.5pt}\color{codekw},
  escapechar={|}, 
}
\begin{lstlisting}[language=python]
def noise_content(content, noise_size):
    """ add noise to content query """
    noise = torch.randn_like(content)*
            torch.norm(content, dim=-1)
    noise_content = (1-noise_size)*content + noise_size*noise
    return noise_content
def noise_pos(pos, noise_size):
    """ add noise to position query """
    bbox = decode(pos)
    noise = torch.randn_like(bbox)
    noise_bbox = bbox + noise_size*noise
    noise_pos = encode(noise_bbox)
    return noise_pos
\end{lstlisting}
\end{algorithm}
\begin{algorithm}[htbp]
\small
\caption{ Training code
}
\label{alg:train}
\definecolor{codeblue}{rgb}{0.25,0.5,0.5}
\definecolor{codegreen}{rgb}{0,0.6,0}
\definecolor{codekw}{RGB}{207,33,46}
\lstset{
  backgroundcolor=\color{white},
  basicstyle=\fontsize{7.5pt}{7.5pt}\ttfamily\selectfont,
  columns=fullflexible,
  breaklines=true,
  captionpos=b,
  commentstyle=\fontsize{7.5pt}{7.5pt}\color{codegreen},
  keywordstyle=\fontsize{7.5pt}{7.5pt}\color{codekw},
  escapechar={|}, 
}
\begin{lstlisting}[language=python]
def train(
          T , #iteration forward times
          perturb_prob, # perturbation probability
          content_ps, #content query perturb size
          pos_ps, #position query perturb size
          supervision_pos, #deep supervision positions
          init_content, #[B, N, C]
          init_pos, #[B, N, 4]
          feats, #[B, L, C, H, W]
          annotations):
  """
  # B: batch
  # N: number of proposal boxes
  """
  solving_path = []
  all_loss = 0
  init_content, init_pos = initialization_layer(feats, init_content, init_pos)
  # supervision for initialization layer 
  all_loss += loss(content, pos, annotations)
  # extra supervision for initialization layer
  # in order to stablize the gradient connection 
  # between refinement layer and initialization layer 
  content, pos = init_content,init_pos
  for i in range(2):
      content, pos = refinement_layer(
                  feats,content, pos)
      all_loss += loss(content, pos, annotations)

  # naive fix-point solving...
  with torch.no_grad():
    content, pos = init_content,init_pos
    solving_path = []
    for i in range(T):
        # refinement aware perturbation
        #   1. add noise to content query
        if torch.rand(1) < perturb_prob :
            content = noise_content(
                        content, content_ps
                        )
        #   2. add noise to pos query
        if torch.rand(1) < perturb_prob :
            pos = noise_pos(pos, pos_ps)
        #   3. project noise 
        content, pos = refinement_layer(
                        feats, content, pos
                        )
        if i in supervision_pos:
            solving_path.append((content, pos))
            
  # deep supervision and gradient construction
  for content, pos in solving_path:
    # refinement aware gradient 
    for i in range(2): 
         content, pos = \ 
         refinement_layer(feats,content, pos)
    all_loss += loss(content, pos, annotations)
  return loss
\end{lstlisting}
\end{algorithm}

\end{document}